\documentclass[journal, twoside]{IEEEtran}

\usepackage{amsmath,amsfonts}
\usepackage{algorithmicx}
\usepackage{algorithm}
\usepackage{array}
\usepackage[caption=false,font=normalsize,labelfont=sf,textfont=sf]{subfig}
\usepackage{textcomp}
\usepackage{stfloats}
\usepackage{url}
\usepackage{verbatim}
\usepackage{graphicx}
\usepackage{cite}

\usepackage{tabularx}
\usepackage{booktabs}
\usepackage{algpseudocode}
\usepackage[dvipsnames]{xcolor}
\usepackage{multirow} 
\usepackage{makecell} 
\usepackage{adjustbox}
\usepackage{anyfontsize}

\newcommand{\figlabel}{Fig.}

\newcommand{\fig}[1]{\figlabel{} \ref{#1}}

\newcommand{\alglabel}{Algorithm}
\newcommand{\alg}[1]{\alglabel{} \ref{#1}}

\newcommand{\tablabel}{Table}
\newcommand{\tablabels}{Tables}
\newcommand{\tab}[1]{\tablabel{} \ref{#1}}
\newcommand{\tabs}[2]{\tablabels{} \ref{#1} and \ref{#2}}
\newcommand{\sectlabel}{Section}
\newcommand{\sect}[1]{\sectlabel{} \ref{#1}}

\newcommand{\sheh}{Shoaib Ehsan}
\newcommand{\shehEMe}{sehsan@essex.ac.uk}
\newcommand{\shehEMs}{s.ehsan@soton.ac.uk}

\newcommand{\sm}{Shakaiba Majeed}
\newcommand{\smEM}{shakaiba33@hanyang.ac.kr}

\newcommand{\mm}{Michael Milford}
\newcommand{\mmEM}{michael.milford@qut.edu.au}

\newcommand{\timur}{Timur Ismagilov}
\newcommand{\timurEM}{ti1e24@soton.ac.uk}

\newcommand{\sdr}{Sarvapali D. Ramchurn}
\newcommand{\sdrEM}{sdr1@soton.ac.uk}

\newcommand{\tvtn}{Nguyen Tan Viet Tuyen }
\newcommand{\tvtnEM}{tuyen.nguyen@soton.ac.uk}

\title{\LARGE 
Joint Multi-Condition Representation Modeling via Matrix Factorization for Visual Place Recognition
}

\author{\timur{}$^{1}$* \IEEEmembership{Graduate Student Member, IEEE}, \sm{}$^{2}$, \mm{}$^{3}$ \IEEEmembership{Senior Member, IEEE},  \tvtn{}$^{1}$ \IEEEmembership{Member, IEEE},
\sdr{}$^{1}$, \sheh{}$^{1,4}$ \IEEEmembership{Senior Member, IEEE} 

\thanks{Manuscript received 30 March 2026; accepted 25 August 2026.}
\thanks{This work was supported in part by U.K. Engineering and Physical Sciences Research Council under Grant EP/Y009800/1 and Grant EP/V00784X/1 and in part by EPSRC-UKRI under Grant UKRI840 (PRIV-LOC).}

\thanks{*Corresponding author.}%
\thanks{$^{1}$\timur{}, \tvtn{}, \sdr{} and \sheh{} are with the School of Electronics and Computer Science, University of Southampton, SO17 1BJ Southampton, U.K. {\tt\footnotesize \timurEM{}, \tvtnEM{}, \sdrEM{}, \shehEMs{}}}%
\thanks{$^{2}$\sm{} is with the  Department of Computer Science and Engineering, Hanyang University,  Seoul 04763, South Korea. {\tt\footnotesize \smEM{}}}%
\thanks{$^{3}$\mm{} is with the QUT Centre for Robotics, School of Electrical
Engineering and Robotics, Brisbane, QLD 4000, Australia. {\tt\footnotesize \mmEM{}}}%
\thanks{$^{4}$\sheh{} is also with the School of Computer Science and Electronic Engineering, University of Essex, Colchester, CO4 3SQ, U.K.  {\tt\footnotesize \shehEMe{}}}%

\thanks{Digital Object Identifier (DOI): see top of this page.}%
}

\begin{document}

\maketitle

\begin{abstract}
We address multi-reference visual place recognition (VPR), where reference sets captured under varying conditions are used to improve localization performance. While deep learning with large-scale training improves robustness, increasing data diversity and model complexity incur extensive computational cost during training and deployment. Descriptor-level fusion via voting or aggregation avoids training, but often targets multi-sensor setups or relies on heuristics with limited gains under appearance and viewpoint change. We propose a training-free, descriptor-agnostic approach that jointly models places using multiple reference descriptors via matrix decomposition into basis representations, enabling projection-based residual matching. We also introduce SotonMV, a benchmark for multi-viewpoint VPR captured over time.
Our method improves Recall@1 by up to $\approx$18\% over conventional single-reference set matching on appearance-change datasets, and matches or outperforms existing multi-reference baselines across most appearance-change settings. For viewpoint-change datasets, our method supports matching when viewpoint is distributed among the references, with the benefit depending on reference set coverage of the query conditions. On unconstrained reference sets, we additionally observe gains of $\approx$5\%, demonstrating strong generalization while remaining computationally lightweight.
\end{abstract}

\begin{IEEEkeywords}
Localization, Recognition, Visual-Based Navigation, Data Sets for Robotic Vision
\end{IEEEkeywords}

\section{Introduction}
\IEEEPARstart{V}{isual} Place Recognition (VPR) is a fundamental capability for robotic systems, autonomous vehicles and Simultaneous Localization and Mapping (SLAM) systems, enabling robust localization and loop-closure detection to support map building and correct global drift. At its core, VPR matches live sensor observations against a database of previously visited locations, relying on topological cues without external positioning systems like GNSS. 
\begin{figure}[!t]
    \centering
\includegraphics[width=0.93\columnwidth{}]{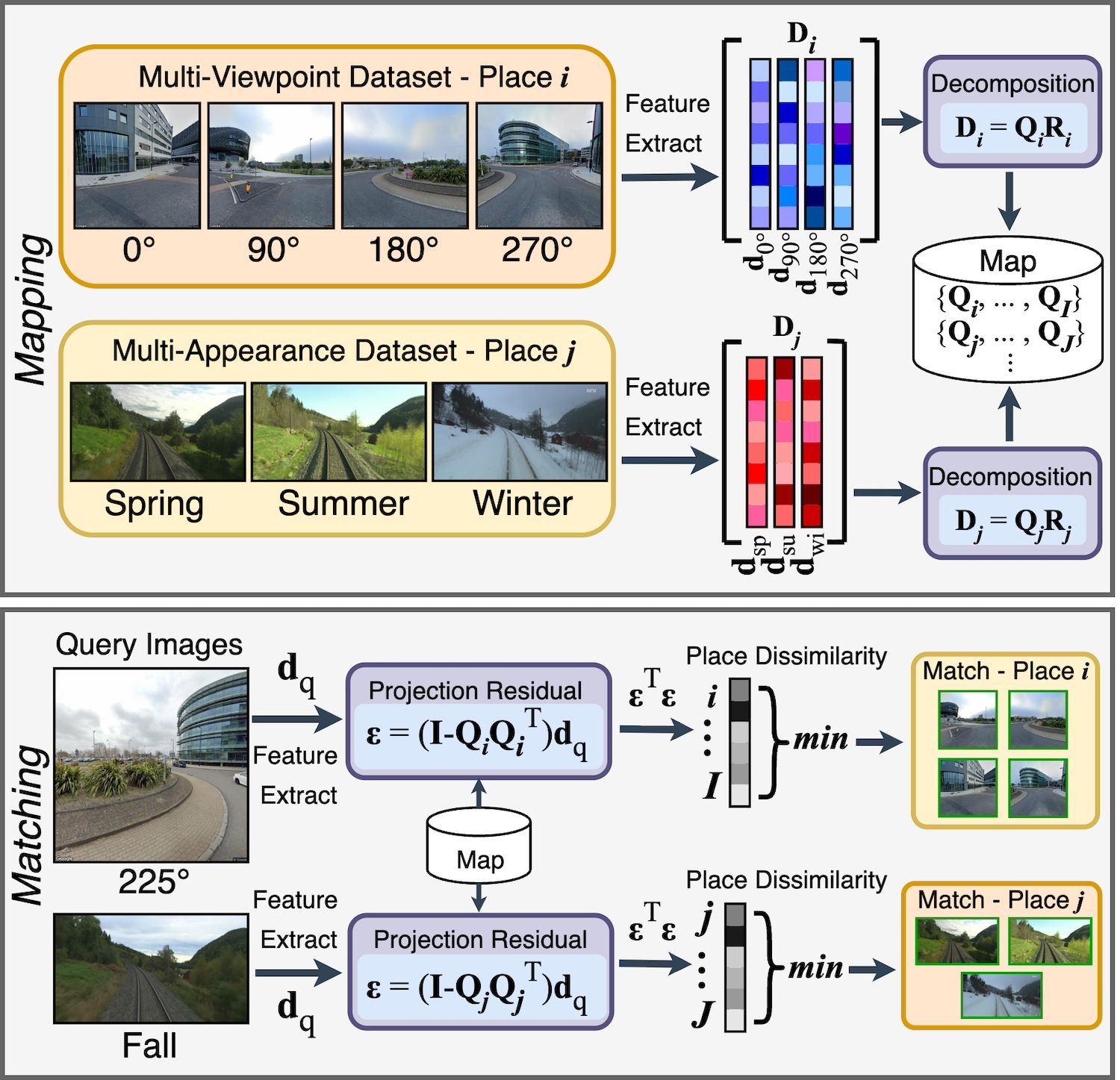}
    \caption{\textit{Top}: Each place is represented by multiple images with different conditions, whose descriptors are stacked and subspace is formed using matrix factorization, here QR decomposition, to construct the map. \textit{Bottom}: At inference, the query is projected and reconstructed using each subspace, with the smallest residual determining the match.}
    \label{fig:Mapping}
    \vspace{-1em}
\end{figure}

VPR is commonly framed as an image retrieval problem, where state-of-the-art (SotA) methods learn to extract distinctive descriptors and perform matching based on similarity. However, long term operation is impacted by perceptual aliasing from appearance variation such as weather, illumination and seasonal change, as well as viewpoint shifts and dynamic scenes, all of which can compromise matching performance.

To improve performance, recent works have proposed more robust feature extractors that utilize pretrained ResNet and DINOv2 backbones, providing robust low-level features transferred from ImageNet training. Recent SotA models then build on such foundation models with effective feature aggregation stages like VLAD, GeM and MixVPR \cite{9577552,ali2023mixvpr,Izquierdo_CVPR_2024_SALAD,Ali-bey_2024_CVPR}, cross-attention mechanisms between images \cite{Ali-bey_2024_CVPR,10656723}, methods for VPR adaption, such as convolutional adapters in backbones \cite{11229956,10656723} as well as specialized loss functions like extensions of Large Margin Cosine Loss \cite{9880209} and combined lateral and frontal classification loss for viewpoint invariance \cite{10377066}. Beyond architecture, data sampling strategies for training have been developed to mine hard recognition examples, improving discriminability and retrieval precision in VPR \cite{Ali-Bey_2022_BMVC,10.1007/978-3-031-73464-9_15}, and improving the effectiveness of standard objectives such as triplet and contrastive loss. Large-scale training on many diverse datasets has shown to be particularly effective \cite{Berton_2025_MegaLoc}, stepping towards unified models capable of broad localization across a variety of domains.

While absolute accuracy has improved, achieving robust global invariance still requires extensive training on highly diverse data, and no current model generalizes universally. This performance is also typically paired with increasing training computational cost and model capacity, with larger models like MegaLoc \cite{Berton_2025_MegaLoc} becoming less practical for edge deployment due to heavier backbones and high-dimensional image representations that dominate memory and latency. Furthermore, VPR in practice often assumes a single reference set, typically collected under favorable daylight conditions and with known camera headings, thereby constraining the range of query viewpoint at inference. Alternatively, strategies exist to improve robustness by providing multiple observations of locations at retrieval time, such as memorizing condition-specific past traversals as separate maps \cite{6224596}, voting across multiple candidate references \cite{9310189}, sequence-to-sequence-based matching \cite{6224623,9842298} and fusion techniques that aggregate information from multiple embeddings or sensors \cite{9588194,9197360,peng2025rangebirdseyeview}. For this work, we focus on the multi-reference setting, where places are represented by multiple observations under varying conditions available at inference time. Despite their efficiency, existing methods treat these views independently or combine them with simple heuristics, leaving approaches that exploit multi-reference information largely under explored. We propose a unified mapping and matching framework that jointly incorporates multiple reference sets at match time. Using matrix decomposition, we construct basis representations that enable inference across diverse conditions, including both appearance and viewpoint shift, without retraining, while remaining lightweight and descriptor-backbone agnostic. Our contributions are as follows:

\begin{itemize}
    \item \textbf{Decomposition-based mapping:} We represent each place by stacking its multi-condition descriptors and applying QR decomposition to obtain an orthonormal basis that captures shared structure across viewpoint and appearance variations, illustrated in \fig{fig:Mapping}. This mapping is training-free, efficient, and supports both pre-computed and on-the-fly updates with minimal overhead. 
    \item \textbf{Projection-based matching:} At retrieval time, queries are projected into each place’s basis, with the residual reconstruction error serving as a similarity metric, selecting the place with the minimum residual.
    \item \textbf{SotonMV:} A multi-view VPR benchmark captured in Southampton, UK, with multiple images per place at regularly sampled orientations. This enables controlled evaluation of multi-view map representations and classic VPR under challenging viewpoint. We also assess if orientation can be inferred from descriptors alone.
    \item \textbf{Comprehensive Evaluation and Analysis:} We evaluate on multi-appearance, viewpoint, and unconstrained reference datasets, consistently improving Recall@1 over single-reference matching and often outperforming multi-reference methods. We analyze computational overhead, as well as the effect of descriptor size and SVD truncation, demonstrating that our approach retains high performance while keeping memory and computation low.
\end{itemize}

In real-world settings, VPR systems, like in SLAM and global localization, must operate under arbitrary viewpoints, traversals, appearance conditions and sparse, uneven map coverage, while remaining computationally efficient. The ability to recall and fuse multiple relevant cues and perspectives across varied conditions into place representations provides a powerful and general foundation for robust localization. This motivates such methods, enabling general applicability under real-world conditions while maintaining low overhead.

The rest of the paper is organized as follows. \sect{sec:relatedwork} reviews related work. \sect{sec:methodology} outlines our proposed framework of mapping and retrieval. SotonMV is detailed in \sect{sec:sotonmv}. \sect{sec:expsetup} covers our experimental setup. Results are presented in \sect{sec:results}. Finally, conclusions, limitations and future work are discussed in \sect{sec:conclusion}.

\section{Related Work}
\label{sec:relatedwork}

\textbf{Visual Place Recognition.}
Retrieval-based VPR typically converts images to distinct feature-level embeddings, which are then matched using distance metrics like cosine distance.
Early methods relied on handcrafted local features like SIFT \cite{10.1023/B:VISI.0000029664.99615.94}, SURF \cite{10.1007/11744023_32} and ORB \cite{6126544}, which are aggregated into global representations using techniques like VLAD \cite{5540039} and Bag-of-Words (BoW) \cite{4509438}. Such methods have since been surpassed with the emergence of deep learning, learning more distinctive and robust representations under complex appearance changes. NetVLAD \cite{7780941} combines VLAD aggregation with a CNN backbone; MixVPR \cite{ali2023mixvpr} uses pre-trained convolutional backbones and channel mixers to effectively aggregate features and create image embeddings. EigenPlaces \cite{10377066} improves viewpoint invariance by grouping images of the same place from different viewpoints, and optimizes them with a lateral and frontal classification loss.
Several methods consider sequences of images during matching \cite{6224623,9842298}, exploiting temporal differences in short image sequences to improve confidence. GeoWarp \cite{9711438}, uses a learned pairwise image homography warping for match re-ranking, improving performance under little visual overlap. 

More recent SotA VPR methods utilize pre-trained transformer backbones. CricaVPR \cite{10656723} uses DINOv2 \cite{oquab2024dinov} as its backbone with convolutional adapters. A cross-image encoder then measures correlation between multiple image representations in a batch for more condition-robust representations. SelaVPR \cite{selavpr} uses a DINOv2 backbone with lightweight adapters, and produces global descriptors for retrieval with dense local features for re-ranking. SelaVPR++ \cite{11229956} builds on SelaVPR, introducing memory-efficient MultiConv adapters, and replaces local feature re-ranking with binary-features for fast retrieval and high-dimensional global descriptors for re-ranking. SALAD \cite{Izquierdo_CVPR_2024_SALAD} uses DINOv2 with an optimal transport-based aggregation extending NetVLAD that considers both feature-to-cluster and cluster-to-feature relations. Bag-of-Queries (BoQ) \cite{Ali-bey_2024_CVPR} supports ResNet50 and DINOv2 backbones. It learns a fixed set of global queries during training to capture universal place-specific attributes. Cross-attention is then used to probe local image features with these queries for selective aggregation into one descriptor. MegaLoc \cite{Berton_2025_MegaLoc} uses DINOv2 with SALAD aggregation, however, it fuses data from multiple diverse datasets for large-scale training to achieve broad generalization with a single model.

Despite advances in deep architectures, VPR is not universally invulnerable to severe appearance changes, limited viewpoint, dynamic scenes and image degradation \cite{10938388}. For example, MegaLoc highlights suboptimal performance on natural scenes like forests and caves, to which Anyloc \cite{keetha2023anyloc} is better suited, using DINOv2 features with VLAD aggregation over domain-specific cluster vocabularies. Nevertheless, universal invariance is largely addressed through network design or large-scale training strategies, as seen with recent models. Training data sampling strategies like GPM \cite{Ali-Bey_2022_BMVC} propose a proxy-based sampling strategy that constructs mini-batches of visually similar places to obtain useful hard negatives for pairwise and triplet loss functions. CliqueMining \cite{10.1007/978-3-031-73464-9_15} improves embedding quality by constructing batches of geographically close images with a graph, forcing the model to learn fine-grained distinctions and better ranking at small distances.

Though effective, large-scale training compute scales with environments \cite{MaloneICCV2025}, particularly if aiming to satisfy global localization and possible perceptual aliasing associated with each domain. Larger models such as AnyLoc and MegaLoc have greater capacity, enabling diverse learning, however, this introduces challenges for deployment on resource-constrained platforms \cite{ijcai2021p603} and TinyML deployment \cite{10284551} where efficiency is crucial and hardware is limited, motivating approaches beyond scaling that minimize required computation. For this work, we explore a novel approach to mapping and matching in VPR, operating at the descriptor level. While existing methods typically improve robustness through architectural changes or retraining, and assume single-image reference sets, we instead model reference places captured under multiple viewpoints and conditions as compact subspace representations via matrix decomposition, extending the robustness of existing methods without modifying the underlying model.

\textbf{Multi-Reference Approaches.} Using multiple observations of the same place is a common strategy for handling environmental variation in VPR. Retrieval-based multi-reference approaches typically target appearance robustness, using season or weather-based datasets such as Nordland \cite{olid2018single}, Oxford RobotCar \cite{RobotCarDatasetIJRR}, or historical GSV imagery \cite{5481932}. These efforts combine descriptors across conditions into single representations via hyperdimensional computing \cite{MaloneICCV2025}, perform sequence search over multi-reference maps \cite{8633431}, or dynamically select relevant reference sets \cite{9310189,7138985}. However, evaluations are mostly restricted to appearance change scenarios, and fusion methods on multiple reference sets are often heuristic and offer limited or inconsistent gains. CricaVPR learns correlations across both appearance and viewpoint conditions, yielding single embedding representations with strong performance, however, this is learned during training, and inference follows single-reference matching. Motivated by works that consider multiple perspectives for place representations, we address mapping and retrieval by jointly decomposing multi-reference descriptors to recover shared basis representations, avoiding explicit fusion or training while maintaining low overhead and versatility with different VPR models.

\begin{figure*}[ht]
    \centering
    \includegraphics[width=0.75\textwidth]{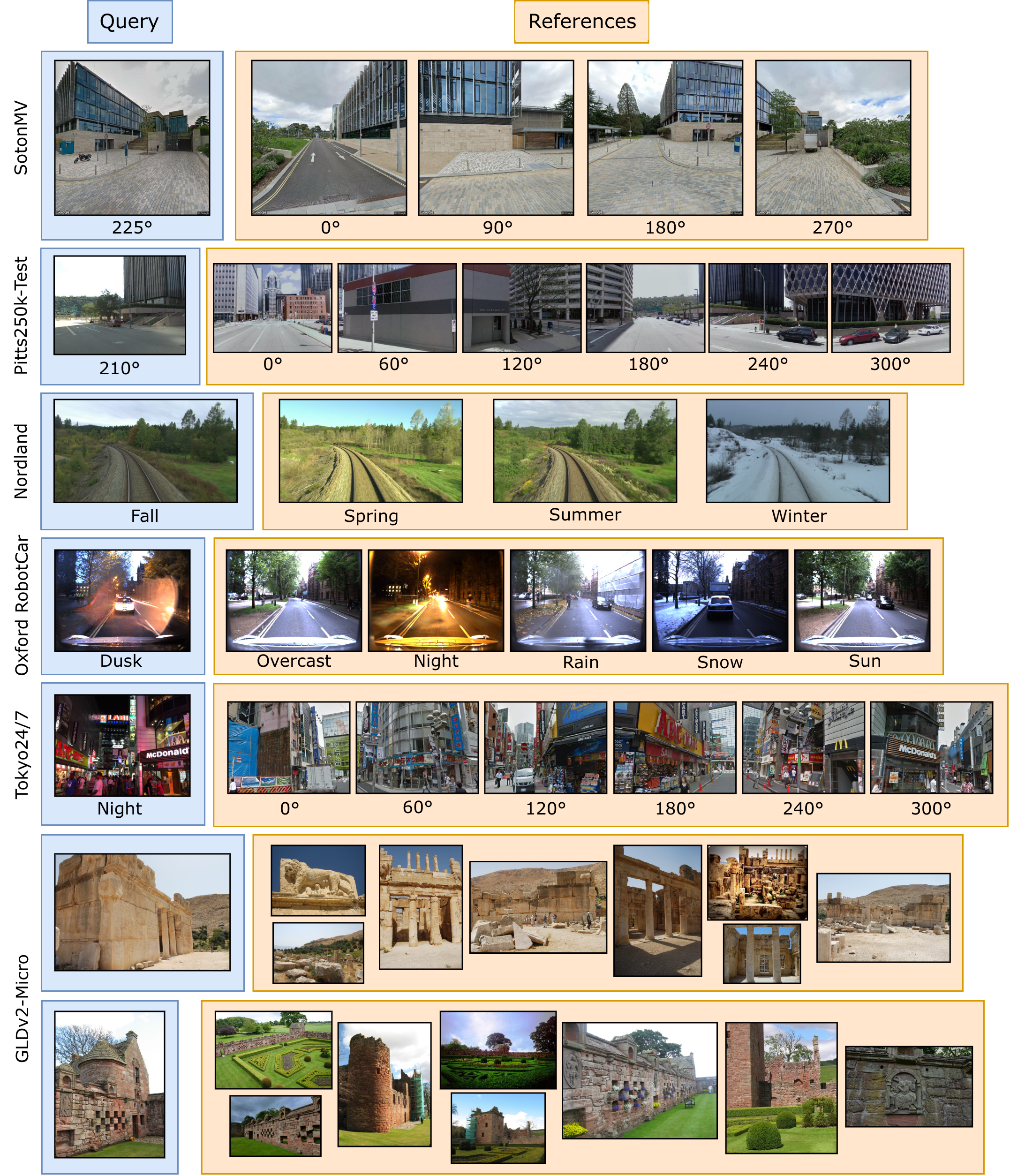}
    \caption{
        Example places from datasets we evaluate. These datasets cover multi-reference sets with varying viewpoints, illumination, seasons and weather. Notably, the GLDv2-micro dataset contains irregular reference imagery with varying conditions, size and channel compositions.
        }
    \label{fig:datasetsamples}
    \vspace*{-1em}
\end{figure*}

Few works explore multi-reference approaches for challenging viewpoint shifts. HOPS \cite{MaloneICCV2025}, which uses hyperdimensional computing to fuse multiple reference descriptors, targets appearance variation rather than viewpoint. \cite{0b2f0681a67c4abf971be1c14ccb69c8} performs sequence-to-single matching by extracting depth keypoints over multiple reference frames to address opposing viewpoints. LoST \cite{garg2018lost_rss} develops semantically-salient keypoint correspondences for spatial verification, enabling place recognition when returning from the opposite direction. Beyond these, a dedicated multi-viewpoint reference aggregation at inference time is largely unexplored, with most viewpoint robustness work like Eigenplaces and CosPlace targeting invariance learned at training time. Furthermore, many VPR datasets like Nordland and St. Lucia \cite{Warren2010} are not organized with places captured at multiple headings, limiting evaluation. SF Landmarks \cite{5995610} contains landmark-centric imagery thereby spanning different bearings, but it is difficult to define consistent "places" across perspectives. GSVCities \cite{ali2022gsv} offers multiple images of the same location with moderate viewpoint shifts while maintaining substantial overlap. Pitts250k \cite{Torii-CVPR2013} and Tokyo24/7 \cite{7850985}, although lacking exact heading metadata, contain imagery spanning full yaw at 30° intervals, enabling experimentation under arbitrary viewpoint in dense urban environments. In this work, we also introduce SotonMV, a Pitts250k-style benchmark comprising places captured at 90° intervals, while also providing repeated acquisitions over 16 years with accurate GPS and relative bearing metadata for experimentation with both multi-viewpoint representations and conventional VPR. 

While not specifically involving multi-reference sets, several works address image-to-panorama \cite{10421857} or panorama-to-panorama \cite{10.1145/3078971.3079033} matching, however, panoramas require specialized capture, suffer stitching artifacts, incur higher memory costs and are incompatible with standard VPR algorithms. Other extensions like FOMP \cite{8461042} fuse omni-directional and multi-sensor data for long term VPR, and aerial-ground collaboration has been widely studied \cite{peng2025rangebirdseyeview,Wang_2025_CVPR}, but both impose extra hardware, training and efficiency costs. In contrast, our method uses single-view cameras with standard VPR embeddings, jointly modeling cross-appearance and cross-viewpoint references at the descriptor level with minimal computational overhead.

\textbf{Subspace Matching.} Subspace matching compares multiple sets of data through their underlying structure and geometric relationships, modeling the variation within a set, unlike independent sample comparisons. Early work on face recognition \cite{139758} represented face images in a "training" set with a low-dimensional linear subspace learned via principal component analysis (PCA). Recognition is then done by comparing the projected vector to stored vectors, using its distance in the subspace, under the assumption that face images are highly correlated and most variation lies in a linear manifold. In recognition tasks, several works build on Mutual Subspace Methods (MSM) for face recognition, action recognition and object tracking, which generally encompasses subspace-to-subspace matching using PCA on sets of images and with no explicit feature learning steps. Constrained MSM (CMSM) \cite{10.5555/1661445.1661626} increases separation between subspace classes made via PCA by removing shared components between classes through eigenvalue analysis. Image-set recognition is then performed by comparing principal angles between subspaces as a distance metric. More recently, Grassmannian Learning MSM (G-LMSM) \cite{10.1016/j.neucom.2022.10.040} extracts feature vectors from each image using a CNN, stacking them into a matrix and applying PCA to convert the set into a subspace. The input subspace is then compared against a set of learned dictionary (class) subspaces using a similarity based on canonical angles. These dictionary subspaces are optimized on the Grassmann manifold via Riemannian gradient descent. Recent work in anomaly detection \cite{Lendering_2026_CVPR} reveals strong few-shot anomaly detection performance using a simple training-free PCA subspace model over frozen DINOv2 patch features. 

Despite extensive use of subspace-based representations in image-set recognition, their application to VPR has not been explored, with existing VPR methods predominantly operating on individual descriptors. We therefore establish a simple subspace-based formulation of VPR as a training-free baseline, representing each reference place as a low-dimensional linear subspace constructed from stacked descriptors via QR decomposition to capture intra-place variation across viewpoints and conditions and performing matching through subspace similarity. Specifically, we use projection-residual matching between individual query descriptors and the reference subspaces, which is equivalent or monotonic to principal-angle formulations (and the Grassmannian distances built on them) in the rank-1 query case. Following subspace methods like MSM and its variants, we avoid mean-centering, keeping subspaces origin-passing and consistent with standard VPR retrieval. We leave alternatives such as CCA, which is defined between two descriptor sets, to future work. Importantly, our approach is agnostic to the feature extractor and introduces no additional training or architectural modifications.

\section{Methodology}
\label{sec:methodology}

\textbf{Motivation.} Matrix factorization is a fundamental tool in linear algebra for decomposing high-dimensional data into structured components that reveal latent relationships \cite{STEINOBRIEN2018790,Lee1999}. They provide orthonormal bases for subspace analysis, simplify the solution of linear systems, and enable dimensionality reduction \cite{doi:10.1137/090771806}. As a result, they are widely used in machine learning \cite{math11122674}, feature extraction \cite{CHAKROBORTY2010693}, recommender systems \cite{ijcai2017p447}, and signal processing \cite{AgarwalMehra2014IJERA}.

QR factorization decomposes a matrix into an orthonormal basis for its column space and is numerically stable, making it well-suited for solving ill-conditioned least squares problems. SVD similarly produces orthonormal bases but orders them by singular values, revealing the rank structure of the matrix and enabling techniques like dimensionality reduction and low-rank approximation. In computer vision, such decompositions have been applied to appearance modeling (e.g. Eigenfaces \cite{139758}), subspace matching techniques for object recognition like MSM \cite{10.5555/1661445.1661626}, rank-revealing feature selection \cite{CHAKROBORTY2010693,MOSLEMI2024124919}, and multimodal fusion using column-pivoted QR \cite{10.1007/s00034-021-01757-y}. Despite their success in other tasks, matrix factorization as a method has not been explored in VPR. We extend this to VPR by stacking multi-condition descriptors of each place and decomposing them to obtain orthonormal basis representations that capture the joint variation across conditions. At matching time, we project the query descriptor onto each subspace, scoring matches via the least squares residual, enabling numerically stable and efficient comparisons. Though deep descriptor spaces are globally non-linear, we assume that with modern feature extractors, for a fixed place, variations in conditions lie on a locally linear manifold that can be approximated by a low-dimensional, linear subspace. We highlight that this formulation does not explicitly model every possible form of variation, but rather captures the shared structure across reference descriptors of the same place to represent the dominant modes of appearance and viewpoint change.

\textbf{Formulation.} We formulate our task in the context of image retrieval, that is, identifying the reference place whose representation most closely matches a given query. Formally, let a single place \(r\) in the reference database \(R\) be captured by \(m\) images, such as at different seasons or viewpoints. For each image \(i\) in place \(r\), we extract an \(n\)-dimensional descriptor like in traditional VPR:
\begin{equation}
    \left\{ \mathbf{d}_r^{(i)} \right\}_{i=1}^{m}, 
    \quad \mathbf{d}_r^{(i)} \in \mathbb{R}^{n}.
\end{equation}
We stack the place descriptors. Since \( n \gg m \), this yields the tall matrix:
\begin{equation}
    \mathbf{D}_r 
    = 
    \left[
        \mathbf{d}_r^{(1)},\,
        \mathbf{d}_r^{(2)},\,
        \dots,\,
        \mathbf{d}_r^{(m)}
    \right]
    \in \mathbb{R}^{n \times m}.
\end{equation}
While cosine distance is commonly used in VPR, we pose matching via a least squares minimization problem:
\begin{equation} \label{eq-obj}
\min_{\mathbf{x} \in \mathbb{R}^m}
\bigl\lVert
\mathbf{d}_q - \mathbf{D}_r\mathbf{x}
\bigr\rVert_2^2,
\end{equation}
which captures how well the query descriptor $\mathbf{d}_q$ can be represented as a linear combination of the reference descriptors in $\mathbf{D}_r$. Our method is summarized in \alg{alg:qr-qrres}. The least squares solution is obtained following the projection theorem. The optimal solution $\hat{\mathbf{d}}_q = \mathbf{D}_r\mathbf{x}_{ls}$ has the property that the projection error is orthogonal to $\text{Range}(\mathbf{D}_r)$, therefore $\hat{\mathbf{d}}_q - \mathbf{d}_q \in \text{Null}(\mathbf{D}_r^T$):
\begin{equation}
\mathbf{D}_r^T(\hat{\mathbf{d}}_q - \mathbf{d}_q) = \mathbf{D}_r^T(\mathbf{D}_r\mathbf{x_{ls}} - \mathbf{d}_q) = 0
\end{equation}
\begin{equation} \label{eq-ls}
\mathbf{x_{ls}} = (\mathbf{D}_r^T\mathbf{D}_r)^{-1}\mathbf{D}_r^T\mathbf{d}_q.
\end{equation}
Now, applying the thin QR decomposition to $\mathbf{D}_r$ results in
\begin{equation}\label{eq-QR}
\mathbf{D}_r = \mathbf{Q}_r\,\mathbf{R}_r,
\end{equation}
with \(\mathbf{Q}_r\in\mathbb{R}^{n\times m}\) forming an orthonormal basis for the column space of $\mathbf{D}_r$, which we store as the place representation, and \(\mathbf{R}_r\in\mathbb{R}^{m\times m}\) being upper triangular and invertible.
Substituting (\ref{eq-QR}) into (\ref{eq-ls}), we get:
\begin{equation} \label{eq-sol}
\mathbf{x_{ls}} = \mathbf{R}_r^{-1}\mathbf{Q}_r^T\mathbf{d}_q.
\end{equation}

\begin{algorithm}[t]
\caption{QR-based Subspace Mapping and Matching}
\label{alg:qr-qrres}
\begin{algorithmic}[1]
\Require Reference places \textit{R}, each represented by \(m\) multi-condition images; query images \textit{Q} and descriptor extractor \(f(\cdot)\).

\State \textcolor{red}{\textbf{Precompute map subspaces:}} 
\ForAll{\(r\in\textit{R}\)} 
    \State \(\mathbf{d}_{r}^{(i)} \leftarrow f(I_{r}^{(i)}) \;\; \forall i=1,\dots,m\)
    \State \(\tilde{\mathbf{d}}_{r}^{(i)} \leftarrow \mathbf{d}_{r}^{(i)} / \|\mathbf{d}_{r}^{(i)}\|_2\) \Comment{$\ell_2$ Normalization}
    \State \(\mathbf{D}_{r} \leftarrow [\,\tilde{\mathbf{d}}_{r}^{(i)}\,]_{i=1}^m\)
    \State \(\mathbf{Q}_{r} \leftarrow \texttt{QR\_Decomposition}(\mathbf{D}_{r})\) \Comment{Or SVD with/without truncation}
    \State Cache \(\mathbf{Q}_{r}\)
\EndFor

\State \textcolor{red}{\textbf{Matching:}} 
\ForAll{\(q\in\textit{Q}\)}
    \State \(\mathbf{d}_q \leftarrow f(q)\)
    \State \(\tilde{\mathbf{d}}_q \leftarrow \mathbf{d}_q / \|\mathbf{d}_q\|_2\) \Comment{$\ell_2$ Normalization}
    \State \(scores \leftarrow [] \)
    \ForAll{\(r\in\textit{R}\)}
        \State \(\mathbf{y}_r \leftarrow \mathbf{Q}_{r}^{\top}\tilde{\mathbf{d}_q}\) \Comment{coordinates in subspace}
        \State \(s_r \leftarrow \|\mathbf{y}_r\|_2\) \Comment{projection magnitude (monotone w.r.t.\ residual)}
        \State \(\texttt{append}(scores, s_r)\)
    \EndFor
    \State Rank \textit{R} by decreasing \(s_r\); output top matches with \(s_r\)
\EndFor
\end{algorithmic}%
\end{algorithm}%

Finally, using (\ref{eq-sol}), we quantify the distance between query and reference as the squared \(\ell_2\)-norm of the residual between query and projected query descriptors:
\begin{equation}
\mathbf{\epsilon}_{q,r} = \|\mathbf{d}_q - \hat{\mathbf{d}_q}\|_2^2 = 
\|\mathbf{d}_q - \mathbf{D}_r\mathbf{x}_{ls}\|_2^2
\end{equation}
\begin{equation}
 = \|\mathbf{d}_q - (\mathbf{Q}_r\mathbf{R}_r)(\mathbf{R}_r^{-1}\mathbf{Q}_r^T\mathbf{d}_q)\|_2^2 = 
\|(\mathbf{I} - \mathbf{Q}_r\mathbf{Q}_r^T)\mathbf{d}_q\|_2^2.
\end{equation}%
Such a decomposition yields a more numerically stable system for least‐squares solutions, with the structure of $\mathbf{R}_r$ reflecting the  relative importance and redundancies among the descriptor columns. Moreover, since $\mathbf{Q}_r$ has orthonormal columns, by the Pythagorean theorem it follows that
\begin{equation}
\|\mathbf{d}_q\|_2^2 = \|\mathbf{Q}_r^T\mathbf{d}_q\|_2^2 + \|(\mathbf{I} - \mathbf{Q}_r\mathbf{Q}_r^T)\mathbf{d}_q\|_2^2.
\end{equation} 
Since all descriptors are $\ell_2$-normalized in VPR, the residual is simplified as %
\begin{equation}
\mathbf{\epsilon}_{q,r} = \|\mathbf{d}_q\|_2^2 - \|\mathbf{Q}_r^T\mathbf{d}_q\|_2^2 = 1 - \|\mathbf{Q}_r^T\mathbf{d}_q\|_2^2. %
\end{equation} %
\begin{figure*}[!t]
  \centering

  \includegraphics[width=0.80\textwidth]{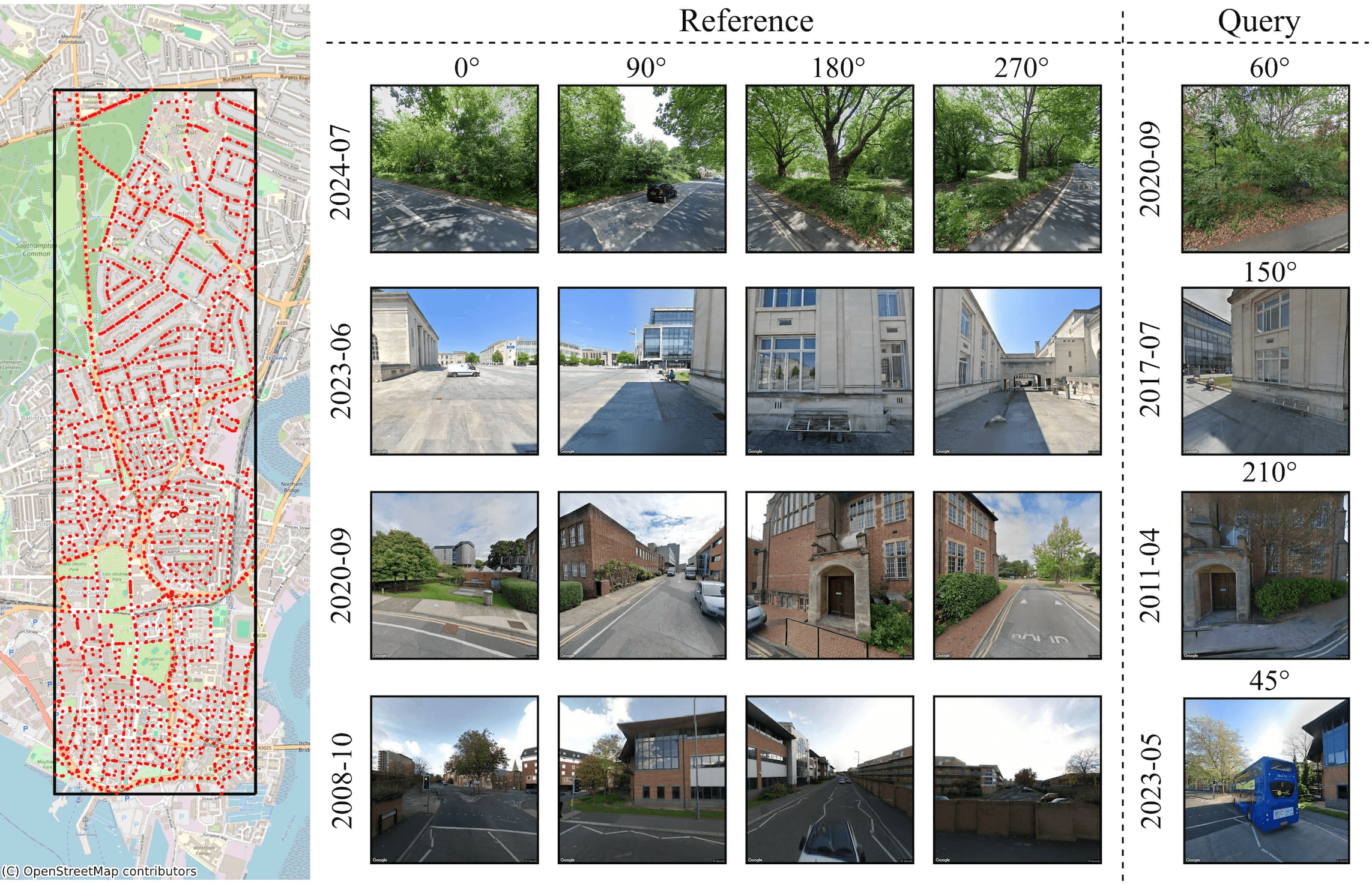}
  \caption{\textit{(Left)} The area considered in Southampton. Each point is a cluster of scenes across multiple dates, with each scene encompassing four images at regularly spaced orientation. \textit{(Right)} Example places from SotonMV. Each place is observed multiple times (appearance change), and captured at four headings (viewpoint change). The query images are taken at intermediate headings, representing VPR under challenging viewpoint with non-trivial spatial overlap.}
  \label{fig:dataset-examples}
\end{figure*}%
\begin{figure*}[!t]
  \centering

  \includegraphics[width=0.93\textwidth]{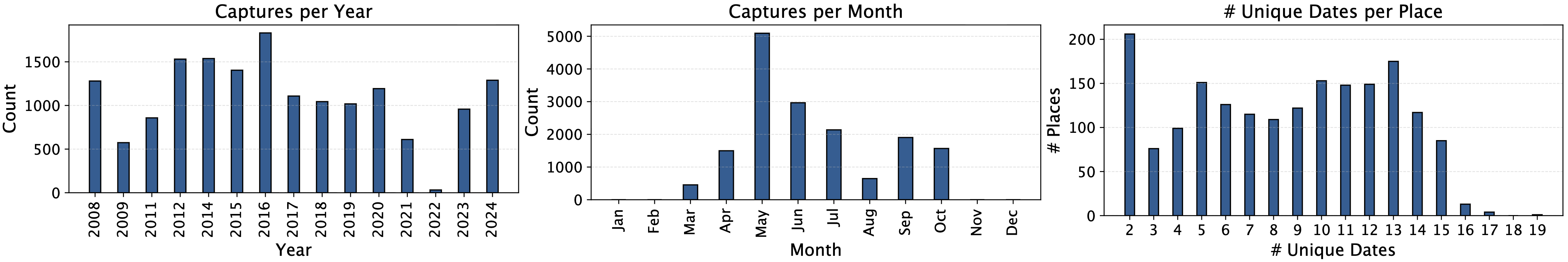}

  \caption{\textit{(Left, Middle)} Distribution of scenes in SotonMV over years and months. \textit{(Right)} Number of unique dates per place. Each place is composed of multiple scenes at different dates, where each scene consists of 4 images at different orientations.}
  \vspace*{-1em}
  \label{fig:dataset-stats}
\end{figure*}%
Therefore, ranking reference places by smallest residual is monotonically equivalent to ranking by largest projection magnitude $\|\mathbf{Q}_r^T\mathbf{d}_q\|_2$, which in practice reduces both time and memory overhead. Retrieval is then performed by ranking reference matches in descending order: 
\begin{equation}
\label{eq:match}
r^{match} = \arg\max_{r\in R} \|\mathbf{Q}_r^T\mathbf{d}_q\|_2.
\end{equation}

\section{SotonMV}
\label{sec:sotonmv}

Google Street View (GSV) \cite{5481932} is a widely used large-scale imagery source in computer vision and image processing \cite{2021LUrbP.21504217B,doi:10.1073/pnas.1700035114,10.1371/journal.pone.0196521}. In VPR, multiple benchmarks \cite{ali2022gsv,Torii-CVPR2013,https://doi.org/10.4218/etrij.17.0116.0034} use GSV due to its coverage, GPS accuracy, and access to historical imagery. We collect a benchmark of multi-view images in Southampton, UK, using the available GSV API. Visualized in \fig{fig:dataset-examples}, the dataset spans approximately $6km^2$ and includes urban and suburban scenes. Places are formed by clustering scenes in a $5m$ radius and discarding those with less than two unique capture dates. The benchmark contains 1849 places with 16268 total scenes, where each scene is composed of four views at relative bearings 0°, 90°, 180°, and 270° with a random pitch between [–5°, 5°] to simulate natural variation, totalling 65072 images. Furthermore, as shown in \fig{fig:dataset-stats} each place is observed at multiple dates between 2008-2024. Although winter-specific conditions are not present due to the limited availability of imagery in these months, the benchmark primarily captures substantial weather, structural and dynamic object changes across time.

The query set includes 300 randomly selected views at intermediate headings 45°, 135°, 225°, and 315° with ±15° offsets, representing challenging viewpoints that may arise under sparse or irregular map coverage. In parallel, 300 randomly selected queries aligned with the reference headings are used to evaluate trivial viewpoint cases. Unlike existing benchmarks with moderate viewpoint shifts, SotonMV provides wide-ranging viewpoints with appearance changes from historical imagery, supported by accurate GPS and relative bearing metadata. Images have 90° FOV, providing complete yaw coverage at each place and isolating the effects of extreme rotation and multi-view representations in VPR, enabling analysis under viewpoint-shift. Additionally, we evaluate on Google Landmarks Dataset v2 (GLDv2) to test performance under real-world irregular and sparse coverage.

\section{Experimental Setup} 
\label{sec:expsetup}

\textbf{Datasets.} We evaluate across datasets that reflect the diversity of challenges faced in real-world deployment. These datasets, shown in \fig{fig:datasetsamples}, include environments with structured variation in season, weather, illumination, or viewpoint, as well as collections where translation, scale, viewpoint and appearance change irregularly. Together, they span urban, suburban and rural settings with challenges from occlusions, dynamic objects and long-term structural changes.

\newcommand{\rot}[1]{\adjustbox{angle=90,valign=l}{#1\hspace{0.3em}}}

\newcolumntype{Y}{>{\centering\arraybackslash}X}

\newlength{\tablew}
\setlength{\tablew}{0.85\textwidth} 
\newlength{\firstcolw}
\setlength{\firstcolw}{3.0cm}       

{
\begin{table*}[!hb]
\small
\centering
\caption{Recall@1 with Nordland retrieval using CNN-based Models.}
\label{tab:nordland}
\setlength{\tabcolsep}{1.2pt}
\renewcommand{\arraystretch}{1.2}

\begin{tabularx}{\tablew}{|p{\firstcolw}|YYYY|YYYY|YYYY|YYYY|}
\hline
\textbf{Queries $\rightarrow$} &
\rot{\textbf{Fall}} & \rot{\textbf{Spring}} & \rot{\textbf{Summer}} & \rot{\textbf{Winter}} &
\rot{\textbf{Fall}} & \rot{\textbf{Spring}} & \rot{\textbf{Summer}} & \rot{\textbf{Winter}} &
\rot{\textbf{Fall}} & \rot{\textbf{Spring}} & \rot{\textbf{Summer}} & \rot{\textbf{Winter}} &
\rot{\textbf{Fall}} & \rot{\textbf{Spring}} & 
\rot{\textbf{Summer}} & \rot{\textbf{Winter}} \\
\cline{1-17}
\textbf{$\downarrow$ References} &
\multicolumn{4}{c|}{\textbf{EigenPlaces (2048D)}} &
\multicolumn{4}{c|}{\textbf{CosPlace (2048D)}} &
\multicolumn{4}{c|}{\textbf{MixVPR (4096D)}} &
\multicolumn{4}{c|}{\textbf{NetVLAD (4096D)}} \\
\hline\hline
\textbf{Fall}   & --- & 88.9 & 93.3 & 62.7 & --- & 85.9 & 91.2 & 58.4 & --- & 95.3 & 95.9 & 81.6 & --- & 47.2 & 69.0 & 17.4 \\

\textbf{Spring} & 80.6 & --- & 73.5 & 67.6 & 76.4 & --- & 68.3 & 68.2 & 88.0 & --- & 83.2 & 89.6 & 42.5 & --- & 37.4 & 21.0 \\

\textbf{Summer} & 94.3 & 82.6 & --- & 59.3 & 92.8 & 79.5 & --- & 53.8 & 95.8 & 91.6 & --- & 78.4 & 69.1 & 42.9 & --- & 18.2 \\

\textbf{Winter} & 51.8 & 70.7 & 47.1 & --- & 46.4 & 66.2 & 41.4 & --- & 70.2 & 86.9 & 64.8 & --- & 14.3 & 22.8 & 13.9 & --- \\

\hline

\textit{Pooling}       & 96.3 & 92.9 & 93.6 & 70.1 & 95.1 & 90.7 & 92.1 & 70.5 & 97.7 & 98.4 & 96.4 & 91.8 & 72.0 & 54.6 & 70.2 & 23.9 \\
\hline
\textit{dMat. Avg.} & 92.8 & 93.3 & 87.3 & 77.9 & 86.5 & 89.1 & 80.9 & 72.2 & 97.1 & 98.3 & 94.8 & 93.1 & 63.8 & 61.6 & 61.9 & \textbf{31.1} \\
\textit{LogSumExp} & 97.1 & 94.8 & 93.5 & \textbf{78.2} & 95.2 & 92.1 & 91.4 & \textbf{74.9} & 98.3 & 98.6 & 96.3 & \textbf{94.0} & 73.8 & 60.0 & 71.7 & 30.5 \\
\textit{Sum Descriptors}   & 96.9 & \textbf{97.2} & 91.6 & 77.5 & 93.9 & 94.9 & 87.7 & 72.3 & 98.4 & 98.8 & 96.1 & 93.7 & 72.0 & \textbf{68.6} & 69.9 & 30.5 \\
\hline
\textit{\textbf{QR-FullVP (Ours)}} & \textbf{98.2} & 97.0 & \textbf{94.6} & 72.8 & \textbf{96.9} & \textbf{95.2} & \textbf{93.1} & 72.2 & \textbf{98.8} & \textbf{98.9} & \textbf{96.7} & 93.8 & \textbf{77.9} & 65.5 & \textbf{76.2} & 27.8 \\
\hline
\end{tabularx}
\end{table*}
} 
\setlength{\tablew}{0.85\textwidth} 
\setlength{\firstcolw}{3.0cm}       

{
\begin{table*}[!ht]
\small
\centering
    \caption{Recall@1 with Nordland retrieval using Transformer-based Models.}
\label{tab:nordland2}
\setlength{\tabcolsep}{1.2pt}
\renewcommand{\arraystretch}{1.2}

\begin{tabularx}{\tablew}{|p{\firstcolw}|YYYY|YYYY|YYYY|YYYY|}
\hline
\textbf{Queries $\rightarrow$} &
\rot{\textbf{Fall}} & \rot{\textbf{Spring}} & \rot{\textbf{Summer}} & \rot{\textbf{Winter}} &
\rot{\textbf{Fall}} & \rot{\textbf{Spring}} & \rot{\textbf{Summer}} & \rot{\textbf{Winter}} &
\rot{\textbf{Fall}} & \rot{\textbf{Spring}} & \rot{\textbf{Summer}} & \rot{\textbf{Winter}} &
\rot{\textbf{Fall}} & \rot{\textbf{Spring}} & 
\rot{\textbf{Summer}} & \rot{\textbf{Winter}} \\
\cline{1-17}
\textbf{$\downarrow$ References} &
\multicolumn{4}{c|}{\textbf{MegaLoc (8448D)}} &
\multicolumn{4}{c|}{\textbf{SALAD (8448D)}} &
\multicolumn{4}{c|}{\textbf{CricaVPR (10752D)}} &
\multicolumn{4}{c|}{\textbf{BoQ (12288D)}} \\
\hline\hline
\textbf{Fall}   & --- & 97.2 & 97.1 & 94.9 & --- & 97.5 & 97.3 & 88.9 & --- & 98.6 & 98.4 & 96.1 & --- & 97.9 & 97.2 & 93.1 \\
\textbf{Spring} & 96.6 & --- & 93.8 & 98.1 & 95.2 & --- & 93.2 & 92.3 & 98.1 & --- & 95.8 & 98.1 & 96.4 & --- & 94.8 & 95.9 \\
\textbf{Summer} & 97.0 & 94.3 & --- & 91.3 & 97.6 & 95.1 & --- & 87.0 & 98.1 & 96.9 & --- & 94.5 & 97.4 & 95.7 & --- & 90.0 \\
\textbf{Winter} & 93.1 & 97.9 & 90.1 & --- & 86.6 & 93.9 & 84.7 & --- & 93.5 & 97.4 & 92.7 & --- & 89.8 & 96.9 & 88.7 & --- \\

\hline

\textit{Pooling} & 99.0 & 99.4 & 97.9 & 98.5 & 99.0 & 99.0 & 97.8 & 94.0 & 99.4 & 99.4 & 98.6 & 99.0 & 99.2 & 99.3 & 97.7 & 97.0 \\
\hline
\textit{dMat. Avg.} & 99.2 & 99.3 & 97.5 & 98.6 & 98.9 & 99.1 & 97.1 & 96.7 & 99.6 & 99.5 & 98.6 & 99.1 & 99.3 & 99.5 & 97.9 & 98.4 \\
\textit{LogSumExp} & \textbf{99.4} & 99.4 & 97.7 & 98.8 & 99.3 & 99.1 & 97.8 & \textbf{97.2} & 99.6 & 99.7 & 98.7 & 99.2 & 99.4 & 99.5 & 97.9 & \textbf{98.7} \\
\textit{Sum Descriptors} & \textbf{99.4} & \textbf{99.5} & 97.8 & 98.8 & 99.4 & \textbf{99.3} & 97.9 & 96.9 & \textbf{99.8} & \textbf{99.8} & 98.9 & 99.2 & \textbf{99.6} & \textbf{99.6} & \textbf{98.2} & 98.4 \\
\hline
\textit{\textbf{QR-FullVP (Ours)}} & \textbf{99.4} & \textbf{99.5} & \textbf{98.1} & \textbf{98.9} & \textbf{99.7} & \textbf{99.3} & \textbf{98.2} & 96.2 & \textbf{99.8} & \textbf{99.8} & \textbf{99.0} & \textbf{99.3} & 99.5 & \textbf{99.6} & \textbf{98.2} & 98.2 \\
\hline
\end{tabularx}
\end{table*}
}
{
\begin{table*}[!ht]
\footnotesize
\centering
\caption{Recall@1 with Oxford RobotCar retrieval using CNN-based Models.}
\label{tab:ORC}

\setlength{\tabcolsep}{0.7pt} 
\renewcommand{\arraystretch}{1.2}
\newcommand{\rowstrut}[2]{\rule{0pt}{#1}\rule[-#2]{0pt}{0pt}}

\begin{tabularx}{\textwidth}{|p{2.5cm}|YYYYYY|YYYYYY|YYYYYY|YYYYYY|}
\hline

\textbf{Queries $\rightarrow$}\rule{0pt}{3.2em} &
\rot{\textbf{Sun}} & \rot{\textbf{Dusk}} & \rot{\textbf{Rain}} & \rot{\textbf{Overcast}} & \rot{\textbf{Night}} & \rot{\textbf{Snow}} &
\rot{\textbf{Sun}} & \rot{\textbf{Dusk}} & \rot{\textbf{Rain}} & \rot{\textbf{Overcast}} & \rot{\textbf{Night}} & \rot{\textbf{Snow}} &
\rot{\textbf{Sun}} & \rot{\textbf{Dusk}} & \rot{\textbf{Rain}} & \rot{\textbf{Overcast}} & \rot{\textbf{Night}} & \rot{\textbf{Snow}} &
\rot{\textbf{Sun}} & \rot{\textbf{Dusk}} & \rot{\textbf{Rain}} & \rot{\textbf{Overcast}} & \rot{\textbf{Night}} & \rot{\textbf{Snow}} \\
\cline{1-25}

\textbf{$\downarrow$ References}\rowstrut{2.2ex}{0.6ex} &
\multicolumn{6}{c|}{\textbf{EigenPlaces (2048D)}} &
\multicolumn{6}{c|}{\textbf{CosPlace (2048D)}} &
\multicolumn{6}{c|}{\textbf{MixVPR (4096D)}} &
\multicolumn{6}{c|}{\textbf{NetVLAD (4096D)}} \\
\hline\hline

\textbf{Sun} & --- & 61.3 & 94.4 & 96.7 & 44.2 & 86.7 & --- & 52.4 & 90.0 & 93.7 & 23.0 & 81.7 & --- & 59.6 & 96.0 & 96.8 & 70.5 & 89.0 & --- & 22.9 & 74.5 & 86.4 & 14.4 & 68.2 \\

\textbf{Dusk} & 57.9 & --- & 54.6 & 58.8 & 35.0 & 62.0 & 49.0 & --- & 42.4 & 49.6 & 17.7 & 50.7 & 51.8 & --- & 49.2 & 52.9 & 55.8 & 57.0 & 16.3 & --- & 14.6 & 17.8 & 16.5 & 25.6 \\

\textbf{Rain} & 95.2 & 57.7 & --- & 94.7 & 44.9 & 84.8 & 89.7 & 48.2 & --- & 89.9 & 25.8 & 78.5 & 95.8 & 55.7 & --- & 95.4 & 68.2 & 85.2 & 78.5 & 23.4 & --- & 80.4 & 15.9 & 57.8 \\

\textbf{Overcast} & 96.3 & 63.6 & 94.2 & --- & 49.2 & 89.7 & 94.0 & 55.3 & 90.6 & --- & 24.4 & 85.4 & 96.8 & 61.8 & 96.0 & --- & 75.4 & 90.5 & 89.3 & 25.4 & 78.4 & --- & 19.2 & 69.1 \\

\textbf{Night} & 62.4 & 47.2 & 60.9 & 66.6 & --- & 60.5 & 41.9 & 32.2 & 38.4 & 45.2 & --- & 37.4 & 69.6 & 58.2 & 70.2 & 74.5 & --- & 69.9 & 14.8 & 17.0 & 11.4 & 15.0 & --- & 17.7 \\

\textbf{Snow} & 90.2 & 69.6 & 88.3 & 93.1 & 50.1 & --- & 84.3 & 61.0 & 82.6 & 87.9 & 26.6 & --- & 91.2 & 67.7 & 89.7 & 93.6 & 72.2 & --- & 64.7 & 34.1 & 55.2 & 68.7 & 14.8 & --- \\
\hline

\textit{Pooling} & 97.7 & 70.7 & 95.7 & 98.4 & 46.1 & 91.7 & 96.0 & 63.2 & 93.5 & 95.6 & 22.7 & 88.1 & 98.1 & 68.9 & 96.8 & 98.1 & 69.2 & 93.1 & 91.4 & 31.4 & 83.5 & 90.5 & 18.4 & 74.0 \\

\hline

\textit{dMat. Avg.} & 97.1 & 73.7 & 94.7 & 97.8 & \textbf{60.0} & 90.8 & 91.4 & 62.8 & 89.1 & 93.1 & 29.0 & 82.8 & 97.6 & \textbf{77.6} & 97.0 & 98.2 & 84.6 & 93.1 & 87.3 & 45.5 & 79.8 & 90.4 & 29.5 & 73.5 \\

\textit{LogSumExp} & 98.7 & 75.6 & 96.7 & 99.1 & 58.2 & 92.6 & 96.7 & \textbf{66.3} & 93.9 & 97.2 & 27.5 & 88.5 & 98.5 & 77.5 & 97.6 & 98.5 & 82.3 & 94.4 & 93.4 & 44.6 & 86.1 & 93.6 & 27.0 & 77.9 \\

\textit{Sum Descriptors} & 98.5 & 75.1 & 97.3 & 99.0 & 58.8 & 93.3 & 96.5 & 65.6 & 94.0 & 97.3 & \textbf{31.2} & 89.0 & 99.0 & 77.3 & 98.3 & 99.2 & \textbf{86.1} & 94.8 & 93.9 & \textbf{46.7} & 87.0 & \textbf{95.4} & \textbf{34.0} & 81.7 \\
\hline

\textit{\textbf{QR-FullVP (Ours)}} & \textbf{98.9} & \textbf{76.2} & \textbf{97.6} & \textbf{99.2} & 53.1 & \textbf{94.4} & \textbf{97.9} & 66.2 & \textbf{95.8} & \textbf{98.4} & 24.9 & \textbf{91.7} & \textbf{99.1} & 77.3 & \textbf{98.4} & \textbf{99.3} & 82.4 & \textbf{95.8} & \textbf{95.2} & 45.4 & \textbf{88.2} & \textbf{95.4} & 28.5 & \textbf{81.9} \\

\hline
\end{tabularx}
\end{table*}
}

{
\begin{table*}[!ht]
\footnotesize
\centering
\caption{Recall@1 with Oxford RobotCar retrieval using Transformer-based Models.}
\label{tab:ORC2}

\setlength{\tabcolsep}{0.7pt} 
\renewcommand{\arraystretch}{1.2}
\newcommand{\rowstrut}[2]{\rule{0pt}{#1}\rule[-#2]{0pt}{0pt}}

\begin{tabularx}{\textwidth}{|p{2.5cm}|YYYYYY|YYYYYY|YYYYYY|YYYYYY|}
\hline

\textbf{Queries $\rightarrow$}\rule{0pt}{3.2em} &
\rot{\textbf{Sun}} & \rot{\textbf{Dusk}} & \rot{\textbf{Rain}} & \rot{\textbf{Overcast}} & \rot{\textbf{Night}} & \rot{\textbf{Snow}} &
\rot{\textbf{Sun}} & \rot{\textbf{Dusk}} & \rot{\textbf{Rain}} & \rot{\textbf{Overcast}} & \rot{\textbf{Night}} & \rot{\textbf{Snow}} &
\rot{\textbf{Sun}} & \rot{\textbf{Dusk}} & \rot{\textbf{Rain}} & \rot{\textbf{Overcast}} & \rot{\textbf{Night}} & \rot{\textbf{Snow}} &
\rot{\textbf{Sun}} & \rot{\textbf{Dusk}} & \rot{\textbf{Rain}} & \rot{\textbf{Overcast}} & \rot{\textbf{Night}} & \rot{\textbf{Snow}} \\
\cline{1-25}

\textbf{$\downarrow$ References}\rowstrut{2.2ex}{0.6ex} &
\multicolumn{6}{c|}{\textbf{MegaLoc (8448D)}} &
\multicolumn{6}{c|}{\textbf{SALAD (8448D)}} &
\multicolumn{6}{c|}{\textbf{CricaVPR (10752D)}} &
\multicolumn{6}{c|}{\textbf{BoQ (12288D)}} \\
\hline\hline

\textbf{Sun} & --- & 69.9 & 94.2 & 95.4 & 77.2 & 84.6 & --- & 72.4 & 95.5 & 96.2 & 83.0 & 89.1 & --- & 71.4 & 96.1 & 97.2 & 79.8 & 89.0 & --- & 77.0 & 97.4 & 97.5 & 88.9 & 90.5 \\

\textbf{Dusk} & 64.7 & --- & 59.8 & 67.2 & 68.3 & 64.6 & 66.7 & --- & 60.6 & 68.1 & 68.9 & 65.4 & 65.2 & --- & 61.7 & 66.0 & 68.8 & 66.4 & 74.4 & --- & 67.3 & 75.3 & 74.5 & 72.2 \\

\textbf{Rain} & 93.6 & 62.6 & --- & 93.0 & 76.2 & 81.1 & 94.8 & 63.4 & --- & 95.1 & 81.8 & 87.1 & 96.4 & 65.6 & --- & 96.2 & 79.5 & 86.0 & 97.0 & 70.5 & --- & 96.8 & 88.8 & 88.5 \\

\textbf{Overcast} & 95.3 & 69.0 & 93.0 & --- & 77.6 & 86.7 & 96.2 & 71.8 & 94.4 & --- & 84.4 & 90.8 & 96.7 & 71.4 & 95.0 & --- & 82.6 & 90.1 & 97.3 & 79.0 & 96.3 & --- & 90.2 & 92.4 \\

\textbf{Night} & 72.5 & 67.7 & 73.1 & 74.0 & --- & 67.9 & 79.9 & 70.1 & 79.1 & 80.2 & --- & 77.8 & 79.8 & 69.9 & 80.4 & 83.0 & --- & 75.7 & 87.4 & 77.0 & 86.1 & 88.3 & --- & 82.9 \\

\textbf{Snow} & 89.2 & 69.2 & 87.3 & 92.2 & 74.6 & --- & 91.0 & 72.9 & 89.5 & 93.3 & 82.7 & --- & 92.4 & 74.3 & 90.5 & 94.8 & 81.0 & --- & 93.5 & 78.6 & 92.2 & 95.6 & 87.5 & --- \\
\hline

\textit{Pooling} & 97.2 & 72.8 & 94.8 & 96.8 & 79.5 & 89.5 & 97.9 & 75.2 & 96.4 & 97.5 & 82.5 & 92.6 & 98.4 & 77.0 & 97.8 & 98.1 & 85.2 & 93.5 & 98.6 & 81.3 & 97.7 & 98.5 & 87.6 & 93.9 \\

\hline

\textit{dMat. Avg.} & 97.1 & 79.1 & 95.6 & 97.5 & \textbf{87.4} & 88.4 & 97.8 & 80.7 & 95.4 & 97.4 & 90.8 & 92.9 & 97.8 & 81.3 & 96.9 & 98.5 & 90.2 & 93.0 & 99.1 & 86.7 & 97.8 & 99.1 & 94.8 & 94.8 \\

\textit{LogSumExp} & 98.1 & 80.3 & 96.2 & 97.9 & 86.9 & 90.5 & 98.8 & 81.3 & 96.7 & 98.2 & 90.2 & 94.4 & 98.8 & 81.2 & 97.5 & 98.6 & 90.0 & 93.7 & 99.2 & 86.8 & 98.0 & 99.1 & 94.5 & 94.9 \\

\textit{Sum Descriptors} & 98.5 & 82.0 & 97.2 & \textbf{98.7} & 88.5 & 91.7 & 99.1 & 81.8 & 97.4 & 98.8 & \textbf{93.1} & 95.0 & \textbf{99.2} & 82.8 & \textbf{98.5} & \textbf{99.4} & \textbf{92.9} & 94.6 & \textbf{99.5} & \textbf{88.2} & 98.5 & \textbf{99.6} & \textbf{96.2} & 96.1 \\
\hline

\textit{\textbf{QR-FullVP (Ours)}} & \textbf{98.6} & \textbf{82.4} & \textbf{97.7} & 98.5 & 87.2 & \textbf{92.7} & \textbf{99.3} & \textbf{82.3} & \textbf{97.7} & \textbf{99.0} & 91.5 & \textbf{96.0} & \textbf{99.2} & \textbf{83.1} & \textbf{98.5} & \textbf{99.4} & 92.1 & \textbf{95.1} & \textbf{99.5} & 88.0 & \textbf{98.7} & 99.5 & 95.2 & \textbf{96.5} \\

\hline
\end{tabularx}
\end{table*}
}

\begin{itemize}
    \item \textit{Nordland}: A $729km$ train route captured across four seasons, enabling evaluation under large-scale appearance change. Each traverse is subsampled to 3975 images, and stationary and tunnel frames are removed.
    \item \textit{Oxford RobotCar}: This benchmark contains multiple traverses collected in Oxford over a year in varied appearance and traffic conditions. We use six different traverses: sun, dusk, rain, overcast, night and snow. We establish direct correspondence between traverses by proximity, obtaining 4054 images in each set with $\approx1m$ interval sub-sampling. Though primarily capturing weather-induced appearance change, traverse intersections and proximity-based sampling occasionally introduce corresponding images with different viewpoints, adding viewpoint change as a secondary challenge.
    \item \textit{SotonMV}: Our benchmark comprises 1849 unique places with four regularly spaced viewpoints per place at relative angles of 0°, 90°, 180° and 270°, captured at multiple dates since 2008. We separately sample 300 queries at challenging transitional viewpoints, excluding their corresponding date imagery from the reference set. SotonMV enables evaluation under extreme rotation along with moderate appearance variation and long-term structural changes.
    \item \textit{Pitts250k-Test}: Contains 3498 unique urban places captured along a traverse in Pittsburgh in similar daylight conditions, from each we sample six images at 60° bearing intervals. We draw 1000 random queries from the provided test split, sampled at intermediate viewpoints. Unlike SotonMV, each place has a single temporal observation but denser angular coverage, with additional challenges from broader spatial sampling of queries.
    \item \textit{Tokyo24/7}: Contains 6332 unique urban places taken in Tokyo in daylight conditions. Like Pitts250k-Test, we sample six images at 60° bearing intervals. Unlike Pitts250k-Test, the query set consists of 105 image triplets sampled at random locations with varying viewpoints, where each triplet includes Day, Sunset, and Night variants and exhibits severe structural changes.
    \item \textit{Google Landmarks Dataset v2 (GLDv2)}: A benchmark with 23294 reference images and 3103 queries. Each place consists of seven to nine ground truth images of varying sizes and formats. Unlike the other datasets where references are captured at fixed viewpoint across season/weather, or with consistent, regularly spaced viewpoints, the references in GLDv2 exhibit vastly irregular changes in viewpoint, translation and appearance (see \fig{fig:datasetsamples}), representing general, real-world collection scenarios like with crowd-sourced data or uncontrolled acquisition over time.
\end{itemize}

\textbf{Approaches.} We benchmark nine well-established VPR models. CosPlace \cite{9880209}, Eigenplaces \cite{10377066}, MixVPR \cite{ali2023mixvpr} and NetVLAD \cite{7780941}, which are commonly used CNN-based models with ResNet50 or VGG16 backbones, and the more recent CricaVPR \cite{10656723}, SALAD \cite{Izquierdo_CVPR_2024_SALAD}, BoQ \cite{Ali-bey_2024_CVPR} and MegaLoc \cite{Berton_2025_MegaLoc}, all of which use the DINOv2 transformer backbone. We further evaluate our approach on SeqVLAD \cite{9842298}, a sequence-to-sequence based VPR method exploiting temporal information. Additionally, we investigate the effect of different descriptor sizes using available pre-trained model variants.

For multi-reference set evaluation, we use pooling as a baseline, where each reference image is considered independently and queries are matched to each descriptor via cosine similarity as done in conventional VPR. We compare distance matrix averaging \cite{9201344}, which averages similarity scores across a place's references into a single ranking; descriptor summation \cite{MaloneICCV2025}, which bundles descriptors via element-wise addition in a hyperdimensional computing framework, and LogSumExp re-ranking, which applies a non-linear score aggregation (log of the sum of exponentials) across the images of top candidates following pooling for candidate re-ranking. Finally, our proposed framework represents each place as a stacked descriptor matrix and derives orthonormal subspaces via QR or SVD decomposition. In this formulation, termed QR-FullVP, each place is modeled as a single compact subspace derived from all available reference views. These basis representations are pre-computed and used for projection-based matching, enabling joint reasoning over multiple references without retraining or architectural changes. On appearance-focused datasets, we also compare conventional single-reference set matching.

\textbf{Metrics.} Performance is measured using Recall@K, assessing the proportion of queries with at least one correct match in the top-K results,  with K=1 being equivalent to the retrieval accuracy. For Nordland and GLDv2, each query has one-to-one ground-truth correspondence. Following the sampling methods of the datasets, Oxford RobotCar and SotonMV use a strict radius tolerance of $5m$, while for Pitts250k and Tokyo24/7, ground-truth matches are defined within a spatial radius of $25m$.

\section{Results}
\label{sec:results}

\begin{table*}[!hb]
  \centering
  \footnotesize
  \setlength{\tabcolsep}{4pt}           
  \renewcommand{\arraystretch}{1.2}     
  \caption{Retrieval performance on SotonMV using CNN-based Models.}
  \label{tab:multiview-soton}
  \begin{tabular}{|l|cccc|cccc|cccc|cccc|}
    \hline
    \textbf{Approach} & \multicolumn{4}{c|}{\textbf{CosPlace (2048D)}} & \multicolumn{4}{c|}{\textbf{Eigenplaces (2048D)}}
    & \multicolumn{4}{c|}{\textbf{MixVPR (4096D)}} & \multicolumn{4}{c|}{\textbf{NetVLAD (4096D)}}\\
    
    \cline{2-17}
    & R@1   & R@5   & R@10  & R@25 
    &  R@1   & R@5   & R@10  & R@25
    & R@1   & R@5   & R@10  & R@25 
    &  R@1   & R@5   & R@10  & R@25\\
    \hline\hline
    
    \textit{Pooling }  & 34.2 & 50.7 & 57.3 & 64.3 & 70.7  & 82.0  & 88.0 & 91.3 & 72.0  & 84.0  & 88.0   & 92.3 & 32.3  & 55.0  & 61.7   & 73.0        \\

    \cline{1-17}

    \textit{dMat. Avg.} & 20.7 & 38.7  
                        & 47.0 & 60.7 & 54.7 & 71.3 & 79.3 &   
                        86.3 & 59.3 & 77.7 & 82.0 &   
                        89.3 & 26.7 & 44.0 & 52.3 & 65.3         
                      \\  
                      \textit{LogSumExp} & 36.7 & 54.0 & 59.0 & 64.3 & 75.7 & 85.7 & 90.0 & 91.3 & 72.7 & 86.3 & 90.3 & 92.3   & 33.7 & 56.3 & 64.3 & 73.0 \\   
    \textit{Sum Descriptors} & 32.3 & 53.0 & 60.7 & 71.3 & 67.3 & 83.7 & 88.0 & 92.3 & 67.3 & 82.3 & 88.3 & 91.3 & 28.7 & 54.7 & 63.7 & 74.7\\ 
    \cline{1-17}
    \textbf{QR-FullVP (Ours)} & \textbf{43.7}  & \textbf{61.7}  & \textbf{67.0}   & \textbf{75.3} & \textbf{80.7}  & \textbf{90.7}  & \textbf{93.0}  & \textbf{94.7} & \textbf{76.3} & \textbf{89.7}  & \textbf{93.0}  & \textbf{95.7} &\textbf{40.3}  & \textbf{62.3}  & \textbf{70.0}  & \textbf{80.0}       \\
   
    \hline

  \end{tabular}
\end{table*}

\begin{table*}[!hb]
  \centering
  \footnotesize
  \setlength{\tabcolsep}{4pt}           
  \renewcommand{\arraystretch}{1.2}     
  \caption{Retrieval performance on SotonMV using Transformer-based Models.}
  \label{tab:multiview-soton2}
  \begin{tabular}{|l|cccc|cccc|cccc|cccc|}
    \hline
   \textbf{Approach} & \multicolumn{4}{c|}{\textbf{MegaLoc (8448D)}} & \multicolumn{4}{c|}{\textbf{SALAD (8448D)}}
    & \multicolumn{4}{c|}{\textbf{CricaVPR (10752D)}} & \multicolumn{4}{c|}{\textbf{BoQ (12288D)}}\\
    
    \cline{2-17}
    & R@1   & R@5   & R@10  & R@25 
    &  R@1   & R@5   & R@10  & R@25
    & R@1   & R@5   & R@10  & R@25 
    &  R@1   & R@5   & R@10  & R@25\\
    \hline\hline
    
    \textit{Pooling } & 78.7 & 91.3 & 94.7 & 96.7 & 76.3 & 88.3 & 93.0 & 96.7 & 57.7 & 74.7 & 81.3 & \textbf{90.0} & 85.7 & 94.3 & 97.3 & 98.7 \\

    \cline{1-17}

    \textit{dMat. Avg.} & 74.0 & 86.3 & 91.7 & 95.3 & 67.3 & 82.0 & 88.3 & 94.0 & 32.0 & 50.3 & 60.3 & 71.3 & 84.0 & 94.7 & 96.3 & 98.3 \\
      \textit{LogSumExp} & 78.3 & 92.0 & 95.3 & 96.7 & 77.7 & 89.0 & 92.3 & 96.7 & 56.0 & 73.3 & 82.0 & \textbf{90.0} & 87.3 & 94.7 & \textbf{98.7} & 98.7 \\
    \textit{Sum Descriptors}  & 76.0 & 89.7 & 93.3 & \textbf{97.3} & 74.0 & 88.0 & 92.7 & 94.3 & 42.7 & 62.3 & 72.7 & 82.0 & 85.0 & 94.3 & 97.7 & 98.7 \\
 
    \cline{1-17}
    \textbf{QR-FullVP (Ours)} & \textbf{81.7} & \textbf{93.7} & \textbf{96.3} & \textbf{97.3} & \textbf{81.3} & \textbf{91.3} &  \textbf{95.3} & \textbf{97.7} & \textbf{60.3} & \textbf{78.0} & \textbf{85.3} & \textbf{90.0} & \textbf{87.7} & \textbf{96.7} & \textbf{98.7} & \textbf{99.0} \\
   
    \hline

  \end{tabular}
\end{table*}

\begin{table*}[!ht]
  \centering
  \footnotesize
  \setlength{\tabcolsep}{4pt}           
  \renewcommand{\arraystretch}{1.2}     
  \caption{Retrieval performance on Pitts250k-Test using CNN-based Models.}
  \label{tab:multiview-pitts}
  \begin{tabular}{|l|cccc|cccc|cccc|cccc|}
    \hline
    \textbf{Approach} & \multicolumn{4}{c|}{\textbf{CosPlace (2048D)}} & \multicolumn{4}{c|}{\textbf{Eigenplaces (2048D)}}
    & \multicolumn{4}{c|}{\textbf{MixVPR (4096D)}} & \multicolumn{4}{c|}{\textbf{NetVLAD (4096D)}}\\
    
    \cline{2-17}
    & R@1   & R@5   & R@10  & R@25 
    &  R@1   & R@5   & R@10  & R@25
    & R@1   & R@5   & R@10  & R@25 
    &  R@1   & R@5   & R@10  & R@25\\
    \hline\hline
    
    \textit{Pooling } & 74.2 & 92.2 & 95.5 & 98.6 & 89.0 & 97.7 & 98.9 & \textbf{99.8} & 89.2  & 96.3  & 98.2   & 99.3 & 55.6 & 76.4 & 84.4 & 91.0     \\

    \cline{1-17}

    \textit{dMat. Avg.} & 54.1 & 77.5 &  86.3 & 96.3 & 73.6 & 89.9 & 93.9 &  98.0 & 77.7 & 90.8 & 95.2 & 98.4 & 37.0 & 59.1 & 71.5 & 84.8       
                      \\  

    \textit{LogSumExp} & 76.4 & 92.9 & 95.8 & 98.6 & 90.7 & 96.5 & 98.4 & \textbf{99.8} & 88.8 & 96.8 & 97.5 & 99.3 & 55.4 & 73.6 & 83.1 & 91.0 \\  
    \textit{Sum Descriptors} & 62.9 & 85.8 & 93.3 & 97.9 & 80.3 & 94.1 & 97.4 & 98.8 & 81.2 & 93.7 & 96.8 & 98.7 & 42.9 & 67.1 & 77.2 & 87.5 \\ 
    
    
    \cline{1-17}
    
    \textbf{QR-FullVP (Ours)}  & \textbf{82.7} & \textbf{95.9} & \textbf{98.0} & \textbf{99.3} & \textbf{93.5} & \textbf{98.6} &  \textbf{99.3} & \textbf{99.8}  & \textbf{91.4} & \textbf{97.9} & \textbf{99.0} & \textbf{99.6} & \textbf{60.3} & \textbf{81.8} & \textbf{88.6} & \textbf{94.0}      \\
   
    \hline

  \end{tabular}
\end{table*}

\begin{table*}[!ht]
  \centering
  \footnotesize
  \setlength{\tabcolsep}{4pt}           
  \renewcommand{\arraystretch}{1.2}     
  \caption{Retrieval performance on Pitts250k-Test (non-default split, as elsewhere - see \sect{sec:expsetup}) using Transformer-based Models.}
  \label{tab:multiview-pitts2}
  \begin{tabular}{|l|cccc|cccc|cccc|cccc|}
    \hline
    \textbf{Approach} & \multicolumn{4}{c|}{\textbf{MegaLoc (8448D)}} & \multicolumn{4}{c|}{\textbf{SALAD (8448D)}}
    & \multicolumn{4}{c|}{\textbf{CricaVPR (10752D)}} & \multicolumn{4}{c|}{\textbf{BoQ (12288D)}}\\
    
    \cline{2-17}
    & R@1   & R@5   & R@10  & R@25 
    &  R@1   & R@5   & R@10  & R@25
    & R@1   & R@5   & R@10  & R@25 
    &  R@1   & R@5   & R@10  & R@25\\
    \hline\hline
    
    \textit{Pooling } & 94.4 & \textbf{99.6} & 99.8 & \textbf{99.9} & 90.3 & 97.9 & 99.0 & 99.4 & 79.1 & 95.8 & 98.0 & \textbf{99.8} & 94.9 & \textbf{99.7} & \textbf{99.9} & \textbf{99.9} \\

    \cline{1-17}

    \textit{dMat. Avg.} & 88.4 & 98.9 & 99.8 & 99.8 & 83.0 & 95.6 & 97.8 & 99.0 & 62.6 & 84.6 & 90.1 & 94.7 & 92.9 & 98.7 & 99.5 & \textbf{99.9} \\ 

    \textit{LogSumExp} & 93.8 & 98.8 & 99.6 & \textbf{99.9} & 91.5 & 97.6 & 98.5 & 99.4 & 81.8 & 94.0 & 97.3 & \textbf{99.8} & 96.2 & 99.0 & 99.4 & \textbf{99.9} \\  
    \textit{Sum Descriptors}  & 90.6 & 99.4 & 99.8 & 99.8 & 87.2 & 97.5 &  98.5 & 99.1 & 68.5 & 88.5 & 93.7 & 97.4 & 94.2 & 99.3 & 99.8 & \textbf{99.9} \\
    
    \cline{1-17}
    
    \textbf{QR-FullVP (Ours)} & \textbf{95.6} & \textbf{99.6} & \textbf{99.9} & \textbf{99.9} & \textbf{93.7} & \textbf{98.6} & \textbf{99.1} & \textbf{99.7} & \textbf{82.3} & \textbf{97.1} & \textbf{98.6} & \textbf{99.8} & \textbf{97.0} & 99.6 & 99.8 & \textbf{99.9} \\
   
    \hline

  \end{tabular}
\end{table*}

\begin{table*}[!ht]
  \centering
  \footnotesize
  \setlength{\tabcolsep}{4pt}           
  \renewcommand{\arraystretch}{1.2}     
  \caption{Retrieval performance on Tokyo24/7 using CNN-based Models.}
  \label{tab:multiview-tokyo}
  \begin{tabular}{|l|cccc|cccc|cccc|cccc|}
    \hline
    \textbf{Approach} & \multicolumn{4}{c|}{\textbf{CosPlace (2048D)}} & \multicolumn{4}{c|}{\textbf{Eigenplaces (2048D)}}
    & \multicolumn{4}{c|}{\textbf{MixVPR (4096D)}} & \multicolumn{4}{c|}{\textbf{NetVLAD (4096D)}}\\
    
    \cline{2-17}
    & R@1   & R@5   & R@10  & R@25 
    &  R@1   & R@5   & R@10  & R@25
    & R@1   & R@5   & R@10  & R@25 
    &  R@1   & R@5   & R@10  & R@25\\
    \hline\hline
    
    \textit{Pooling } & 53.3 & 69.2 & 75.2 & 84.4 & 71.8 & 83.2 & 88.3 & 93.7 & 75.9 & \textbf{87.3} & \textbf{90.8} & \textbf{93.0} & 22.9 & 40.6 & 54.6 & 63.8 \\

    \cline{1-17}

    \textit{dMat. Avg.} & 23.2 & 47.0 & 58.1 & 71.1 & 39.4 & 57.8 & 65.1 & 77.5 & 42.5 & 65.4 & 70.5 & 81.0 & 16.2 & 28.3 & 36.2 & 49.8 \\

    \textit{LogSumExp} & 53.0 & 69.8 & 74.9 & 84.4 & 70.5 & 83.2 & 87.9 & 93.7 & 71.4 & 83.2 & 88.9 & \textbf{93.0} & 22.5 & 37.5 & 49.8 & 63.8 \\
    
    \textit{Sum Descriptors} & 37.5 & 60.6 & 66.4 & 78.4 & 47.3 & 68.3 & 76.2 & 82.9 & 51.4 & 70.8 & 77.8 & 83.8 & 19.1 & 34.3 & 40.0 & 58.4 \\
    
    \cline{1-17}
    
    \textbf{QR-FullVP (Ours)} & \textbf{55.9} & \textbf{73.7} & \textbf{81.3} & \textbf{89.5} & \textbf{75.6} & \textbf{85.7} & \textbf{90.8} & \textbf{94.9} & \textbf{76.5} & 87.0 & 89.2 & 92.4 & \textbf{25.4} & \textbf{44.8} & \textbf{56.5} & \textbf{67.0} \\
    
    \hline

  \end{tabular}
\end{table*}

\begin{table*}[!ht]
  \centering
  \footnotesize
  \setlength{\tabcolsep}{4pt}           
  \renewcommand{\arraystretch}{1.2}     
  \caption{Retrieval performance on Tokyo24/7 (non-default split, as elsewhere - see \sect{sec:expsetup}) using Transformer-based Models.}
  \label{tab:multiview-tokyo2}
  \begin{tabular}{|l|cccc|cccc|cccc|cccc|}
    \hline
    \textbf{Approach} & \multicolumn{4}{c|}{\textbf{MegaLoc (8448D)}} & \multicolumn{4}{c|}{\textbf{SALAD (8448D)}}
    & \multicolumn{4}{c|}{\textbf{CricaVPR (10752D)}} & \multicolumn{4}{c|}{\textbf{BoQ (12288D)}}\\
    
    \cline{2-17}
    & R@1   & R@5   & R@10  & R@25 
    &  R@1   & R@5   & R@10  & R@25
    & R@1   & R@5   & R@10  & R@25 
    &  R@1   & R@5   & R@10  & R@25\\
    \hline\hline
    
    \textit{Pooling } & \textbf{95.9} & 98.4 & 99.1 & \textbf{99.7} & 88.9 & 96.5 & 97.1 & \textbf{99.1} & \textbf{80.0} & \textbf{90.2} & \textbf{93.3} & \textbf{95.6} & 94.0 & 97.8 & 98.4 & \textbf{98.7} \\

    \cline{1-17}

    \textit{dMat. Avg.} & 85.1 & 94.3 & 98.4 & 99.4 & 61.3 & 83.5 & 89.2 & 93.7 & 32.4 & 53.7 & 64.8 & 75.9 & 87.3 & 97.1 & \textbf{98.7} & \textbf{98.7} \\

    \textit{LogSumExp} & 94.0 & 98.4 & 99.1 & \textbf{99.7} & 87.9 & 94.6 & 97.1 & \textbf{99.1} & 71.8 & 86.4 & 90.8 & \textbf{95.6} & 94.3 & \textbf{98.1} & 98.4 & \textbf{98.7} \\
    
    \textit{Sum Descriptors} & 88.3 & 96.2 & 98.7 & 99.1 & 74.6 & 90.2 & 93.3 & 97.1 & 39.7 & 61.3 & 72.1 & 81.0 & 89.8 & 97.1 & \textbf{98.7} & \textbf{98.7} \\
    
    \cline{1-17}
    
    \textbf{QR-FullVP (Ours)} & 94.8 & \textbf{99.1} & \textbf{99.7} & \textbf{99.7} & \textbf{91.4} & \textbf{96.8} & \textbf{97.5} & 98.4 & 76.5 & 88.9 & 92.1 & 94.9 & \textbf{94.6} & 97.8 & 98.4 & \textbf{98.}7 \\
    
    \hline

  \end{tabular}
\end{table*}

\begin{table*}[!ht]
  \centering
  \footnotesize
  \setlength{\tabcolsep}{4pt}           
  \renewcommand{\arraystretch}{1.2}     
  \caption{Retrieval performance on the GLDv2-micro dataset using CNN-based Models.}
  \label{tab:gldv2}
  \begin{tabular}{|l|cccc|cccc|cccc|cccc|}
    \hline
    \textbf{Approach} & \multicolumn{4}{c|}{\textbf{Eigenplaces (2048D)}} & \multicolumn{4}{c|}{\textbf{CosPlace (2048D)}}
    & \multicolumn{4}{c|}{\textbf{MixVPR (4096D)}} & \multicolumn{4}{c|}{\textbf{NetVLAD (4096D)}}\\
    
    \cline{2-17}
    & R@1   & R@5   & R@10  & R@25 
    &  R@1   & R@5   & R@10  & R@25
    & R@1   & R@5   & R@10  & R@25 
    &  R@1   & R@5   & R@10  & R@25\\
    \hline\hline
    
    \textit{Pooling } & 53.6 & 64.7 & 69.6 & 75.3 & 48.1 & 60.0 & 65.2 & 72.1 & 59.7 & 71.0 & 75.4 & 80.9 & 41.1 & 54.9 & 61.0 & 69.6  \\

    \cline{1-17}

    \textit{dMat. Avg.} & 35.2 & 51.3 & 58.0 & 66.7 & 28.4 & 45.0 & 52.8 & 62.4 & 45.2 & 60.6 & 66.7 & 74.6 & 29.1 & 46.6 & 53.9 & 64.7 \\  

    \textit{LogSumExp} & 52.6 & 65.5 & 70.7 & 75.3 & 47.2 & 60.9 & 67.1 & 72.1 & 57.1 & 71.0 & 76.5 & 80.9 & 41.1& 56.5 & 63.8 & 69.6\\ 

    \textit{Sum Descriptors} & 46.4 & 61.7 & 67.3 & 74.2 & 39.5 & 56.3 & 62.4 & 70.7 & 54.8 & 67.9 & 73.1 & 79.3 & 40.4 & 55.8 & 62.2 & 71.1\\ 
    
    \cline{1-17}
    
    \textbf{QR-FullVP (Ours)}  & \textbf{58.8} & \textbf{70.4} & \textbf{74.9} & \textbf{80.3} & \textbf{53.3} & \textbf{66.0} & \textbf{71.0} & \textbf{77.7} & \textbf{64.9} & \textbf{76.4} & \textbf{79.8} & \textbf{84.7} & \textbf{48.7} & \textbf{62.7} & \textbf{68.4} & \textbf{75.4}    \\
   
    \hline

  \end{tabular}
  \vspace*{-1em}
\end{table*}

\begin{table*}[!ht]
  \centering
  \footnotesize
  \setlength{\tabcolsep}{4pt}           
  \renewcommand{\arraystretch}{1.2}     
  \caption{Retrieval performance on the GLDv2-micro dataset using Transformer-based Models.}
  \label{tab:gldv2_2}
  \begin{tabular}{|l|cccc|cccc|cccc|cccc|}
    \hline
    \textbf{Approach} & \multicolumn{4}{c|}{\textbf{MegaLoc (8448D)}} & \multicolumn{4}{c|}{\textbf{SALAD (8448D)}}
    & \multicolumn{4}{c|}{\textbf{CricaVPR (10752D)}} & \multicolumn{4}{c|}{\textbf{BoQ (12288D)}}\\
    
    \cline{2-17}
    & R@1   & R@5   & R@10  & R@25 
    &  R@1   & R@5   & R@10  & R@25
    & R@1   & R@5   & R@10  & R@25 
    &  R@1   & R@5   & R@10  & R@25\\
    \hline\hline
    
    \textit{Pooling} & 86.8 & 92.3 & 94.0 & 95.9 & 71.8 & 82.2 & 85.3 & 89.1 & 70.2 & 81.4 & 84.8 & 88.8 & 74.2 & 83.6 & 86.5 & 89.9 \\

    \cline{1-17}

    \textit{dMat. Avg.} & 80.5 & 89.4 & 92.6 & 94.8 & 59.1 & 74.3 & 79.0 & 85.2 & 52.1 & 67.4 & 73.3 & 80.1 & 67.2 & 80.2 & 84.4 & 88.4 \\

    \textit{LogSumExp} & 85.1 & 92.7 & 94.6 & 95.9 & 70.2 & 82.5 & 86.5 & 89.1 & 66.4 & 79.6 & 84.5 & 88.8 & 71.7 & 83.1 & 86.9 & 89.9 \\

    \textit{Sum Descriptors} & 84.6 & 92.3 & 94.4 & 95.8 & 68.7 & 79.9 & 84.3 & 88.9 & 65.9 & 79.3 & 83.7 & 88.2 & 71.9 & 83.4 & 86.7 & 90.2 \\
    
    \cline{1-17}
    
    \textbf{QR-FullVP (Ours)} & \textbf{89.4} & \textbf{94.3} &\textbf{ 95.7} & \textbf{97.0} & \textbf{76.8} & \textbf{86.2} & \textbf{89.1} & \textbf{92.3} & \textbf{76.2} & \textbf{86.4} & \textbf{88.9} & \textbf{92.8} & \textbf{77.5} & \textbf{86.7} & \textbf{89.4} & \textbf{92.4} \\
   
    \hline

  \end{tabular}
  \vspace*{-1em}
\end{table*}

\begin{table}[t]
\centering
\footnotesize
\caption{Retrieval performance using \textbf{SeqVLAD} across datasets.}
\label{tab:seqvlad}

\setlength{\tabcolsep}{6.2pt}
\renewcommand{\arraystretch}{1.15}

\begin{tabular}{|p{2.2cm}|c c c c|}
\hline
\multicolumn{5}{|c|}{\textbf{Nordland}} \\
\hline

\textbf{Queries $\rightarrow$}
& \rot{\textbf{Fall}}
& \rot{\textbf{Spring}}
& \rot{\textbf{Summer}}
& \rot{\textbf{Winter}} \\

\cline{1-5}

\textbf{$\downarrow$ References}
& \multicolumn{4}{c|}{\textbf{SeqVLAD (24576D)}} \\

\hline\hline

\textbf{Fall}   & --- & 85.7 & 95.0 & 68.4 \\
\textbf{Spring} & 78.9 & --- & 80.0 & 77.3 \\
\textbf{Summer} & 92.2 & 83.6 & --- & 69.4 \\
\textbf{Winter} & 57.9 & 74.2 & 60.9 & --- \\

\hline

\textit{Pooling} & 94.5 & 93.4 & 95.7 & 82.2 \\

\hline

\textit{dMat. Avg.}      & 89.5 & 93.9 & 92.1 & 86.1 \\
\textit{LogSumExp}       & 94.6 & 95.4 & 96.1 & 87.9 \\
\textit{Sum Descriptors} & 93.6 & 97.2 & 95.0 & \textbf{88.6} \\

\hline

\textit{\textbf{QR-FullVP (Ours)}} & \textbf{96.5} & \textbf{97.5} & \textbf{96.9} & 87.1 \\

\hline
\end{tabular}

\vspace{0.9em}

\setlength{\tabcolsep}{3.2pt}

\begin{tabular}{|p{2.3cm}|c c c c c c|}
\hline
\multicolumn{7}{|c|}{\textbf{Oxford RobotCar}} \\
\hline

\textbf{Queries $\rightarrow$}
& \rot{\textbf{Sun}}
& \rot{\textbf{Dusk}}
& \rot{\textbf{Rain}}
& \rot{\textbf{Overcast}}
& \rot{\textbf{Night}}
& \rot{\textbf{Snow}} \\

\cline{1-7}

\textbf{$\downarrow$ References}
& \multicolumn{6}{c|}{\textbf{SeqVLAD (24576D)}} \\

\hline\hline

\textbf{Sun} & --- & 50.2 & 86.2 & 89.6 & 40.2 & 72.2 \\
\textbf{Dusk} & 42.7 & --- & 41.1 & 46.1 & 39.8 & 48.8 \\
\textbf{Rain} & 85.2 & 45.9 & --- & 85.1 & 40.6 & 67.4 \\
\textbf{Overcast} & 90.6 & 51.9 & 87.6 & --- & 41.1 & 75.9 \\
\textbf{Night} & 38.7 & 43.6 & 39.2 & 43.4 & --- & 37.0 \\
\textbf{Snow} & 77.8 & 59.9 & 77.6 & 82.8 & 40.2 & --- \\

\hline

\textit{Pooling} & 93.9 & 63.6 & 91.8 & 93.6 & 46.1 & 80.6 \\

\hline

\textit{dMat. Avg.} & 91.9 & 67.0 & 89.5 & 92.8 & 56.3 & 80.7 \\
\textit{LogSumExp} & 95.6 & 69.2 & 92.7 & 94.9 & 56.0 & 83.4 \\
\textit{Sum Descriptors} & 96.0 & 72.2 & 94.6 & 96.4 & \textbf{61.4} & 87.6 \\

\hline

\textit{\textbf{QR-FullVP (Ours)}} & \textbf{97.1} & \textbf{73.8} & \textbf{95.6} & \textbf{96.7} & 59.8 & \textbf{88.5} \\

\hline
\end{tabular}

\vspace{0.9em}

\setlength{\tabcolsep}{4pt}

\begin{tabular}{|l|c c c c|}
\hline
\multicolumn{5}{|c|}{\textbf{Pitts250k-Test -- SeqVLAD (24576D)}} \\
\hline

\textbf{Approach}
& R@1
& R@5
& R@10
& R@25 \\

\hline\hline

\textit{Pooling} & 45.4 & 70.2 & 80.2 & 89.1 \\

\hline

\textit{dMat. Avg.}      & 22.6 & 39.8 & 48.9 & 66.5 \\
\textit{LogSumExp}       & 39.4 & 57.9 & 70.3 & 89.1 \\
\textit{Sum Descriptors} & 31.0 & 50.5 & 62.2 & 73.2 \\

\hline

\textit{\textbf{QR-FullVP (Ours)}} & \textbf{52.4} & \textbf{73.8} & \textbf{82.0} & \textbf{91.4} \\

\hline
\end{tabular}
\vspace{-1em}
\end{table}

\subsection{Multi-Appearance Datasets}

\begin{figure*}[!hb]
    \centering
    \includegraphics[width=0.91\textwidth]{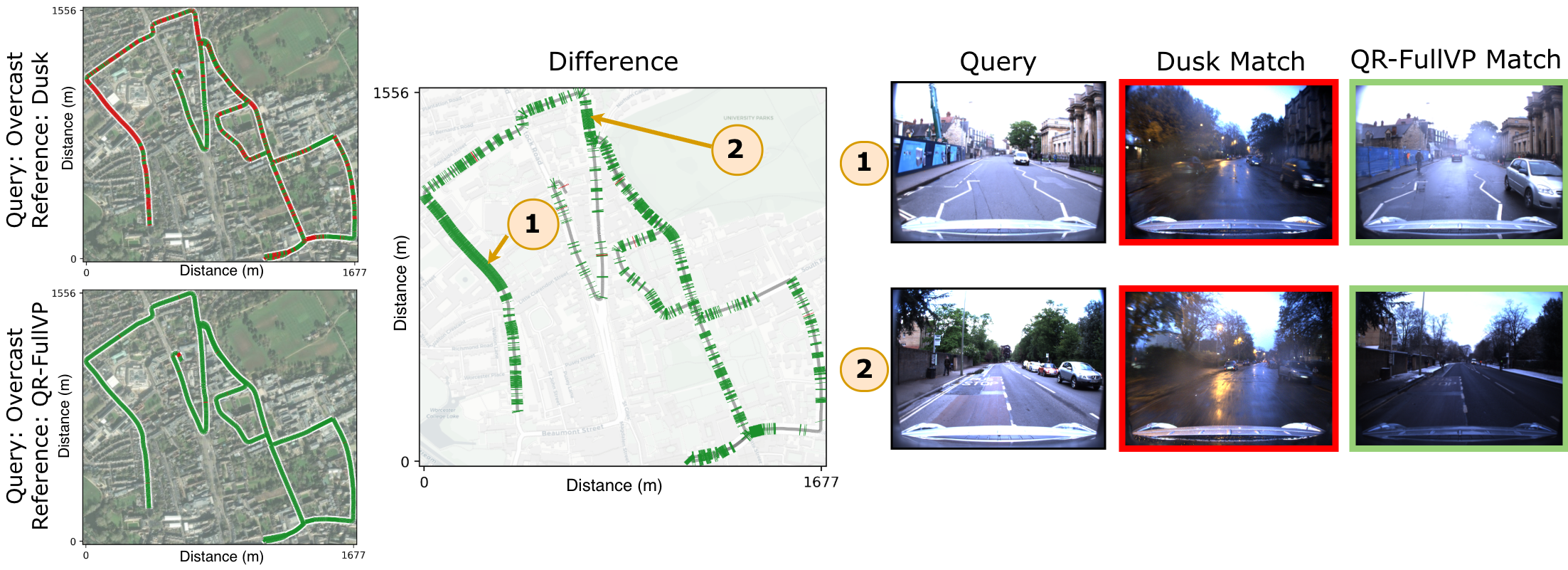}
    \caption{
        \textbf{Retrieval comparison on Oxford RobotCar (overcast query) using Eigenplaces.} Left shows retrieval outcomes for single reference matching (dusk reference) and QR-FullVP (all references), with green indicating correct and red incorrect top-1 matches. The difference map (center) highlights where QR-FullVP succeeds over single-reference (green), the reverse (red) or both agree (grey). As expected, QR-FullVP resolves ambiguity across large portions of the traverse. (Retrieved images for QR-FullVP show a randomly sampled image from the matched place's reference collection).}
    \label{fig:map1}
    \vspace*{-1em}
\end{figure*}

Performance with multi-appearance reference sets are shown in Tables \ref{tab:nordland}, \ref{tab:nordland2}, \ref{tab:ORC}, and \ref{tab:ORC2}, for Nordland and Oxford RobotCar. Compared to single-reference matching, our QR decomposition-based approach outperforms in nearly all (79/80) cases, with improvements across all VPR methods. Overall, we observe gains of up to $\approx$18\% Recall@1 compared to the top performing single-reference set, with significant gains across all seasons and weather conditions, as well as challenging cases like CricaVPR with Oxford-RobotCar dusk queries achieving an improvement of 8.8\%. Qualitative results on Oxford-RobotCar (\fig{fig:map1}) illustrate this, using overcast and dusk traverses as query and reference, respectively. QR-FullVP resolves ambiguity across large portions of the traverse compared to single-image matching. The use of multiple reference sets to capture appearance variation shows to be highly effective, with our framework providing a consistent approach for multi-reference VPR.

\begin{figure*}[!hb]
    \centering
    \includegraphics[width=0.91\textwidth]{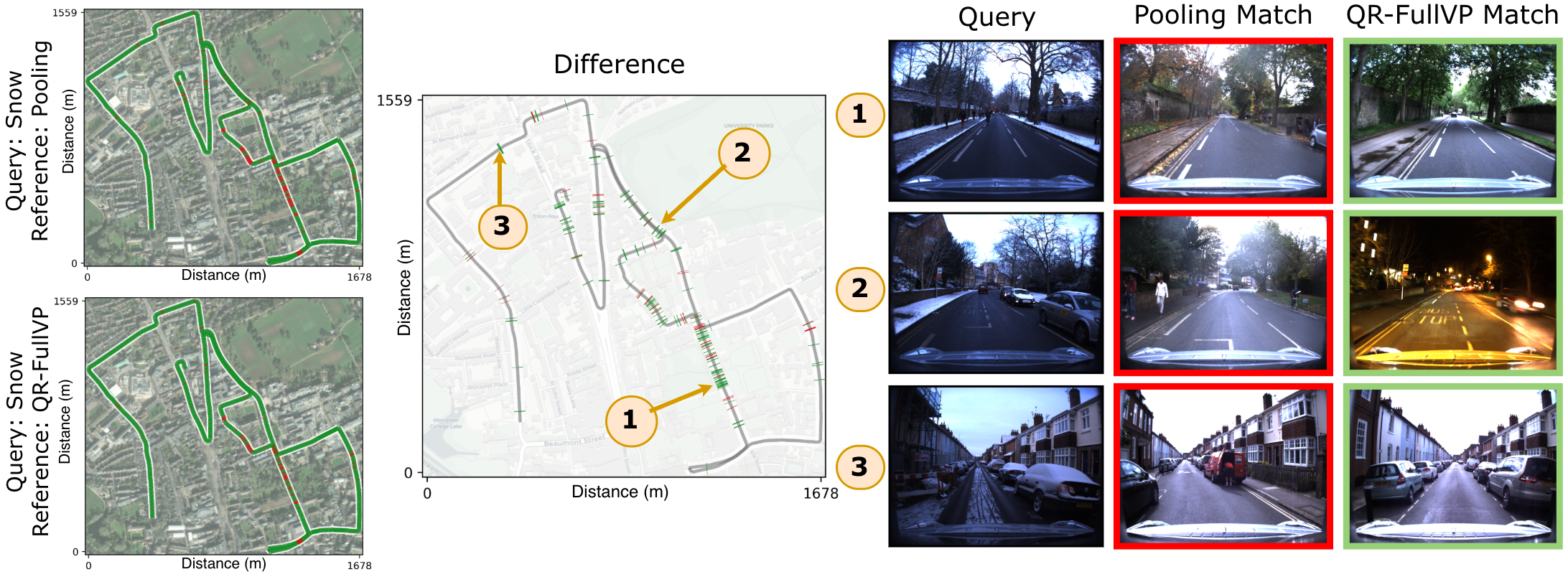}
    \caption{
       \textbf{Retrieval comparison on Oxford RobotCar (snow query) using Eigenplaces.} Left shows retrieval outcomes for reference pooling and QR-FullVP, with green indicating correct and red incorrect top-1 matches. The difference map (center) highlights where QR-FullVP succeeds over pooling (green), the reverse (red), or both agree (grey). Overall, QR-FullVP outperforms pooling, and qualitative examples suggest this is often a fine disambiguation, such as the third example where the pooling match falls just outside the ground truth tolerance while QR-FullVP recovers the correct place. (Retrieved images for QR-FullVP show a randomly sampled image from the matched place's reference collection).}
    \label{fig:map2}
    \vspace*{-1em}
\end{figure*}

With respect to the multi-reference set approaches, our method performs competitively or surpasses others in the majority of cases (23/32 Nordland, 34/48 ORC), while outperforming pooling in all cases with gains up to $\approx$14\% Recall@1, highlighting the benefit of multi-reference representations, rather than treating them individually. As shown in \fig{fig:map2}, QR-FullVP generally outperforms pooling across portions of the traverse under snow conditions, and we find there are frequent cases of fine-grained disambiguation, with QR-FullVP retrieving a more precise match.

\begin{figure*}[!ht]
    \centering
    \includegraphics[width=0.91\textwidth]{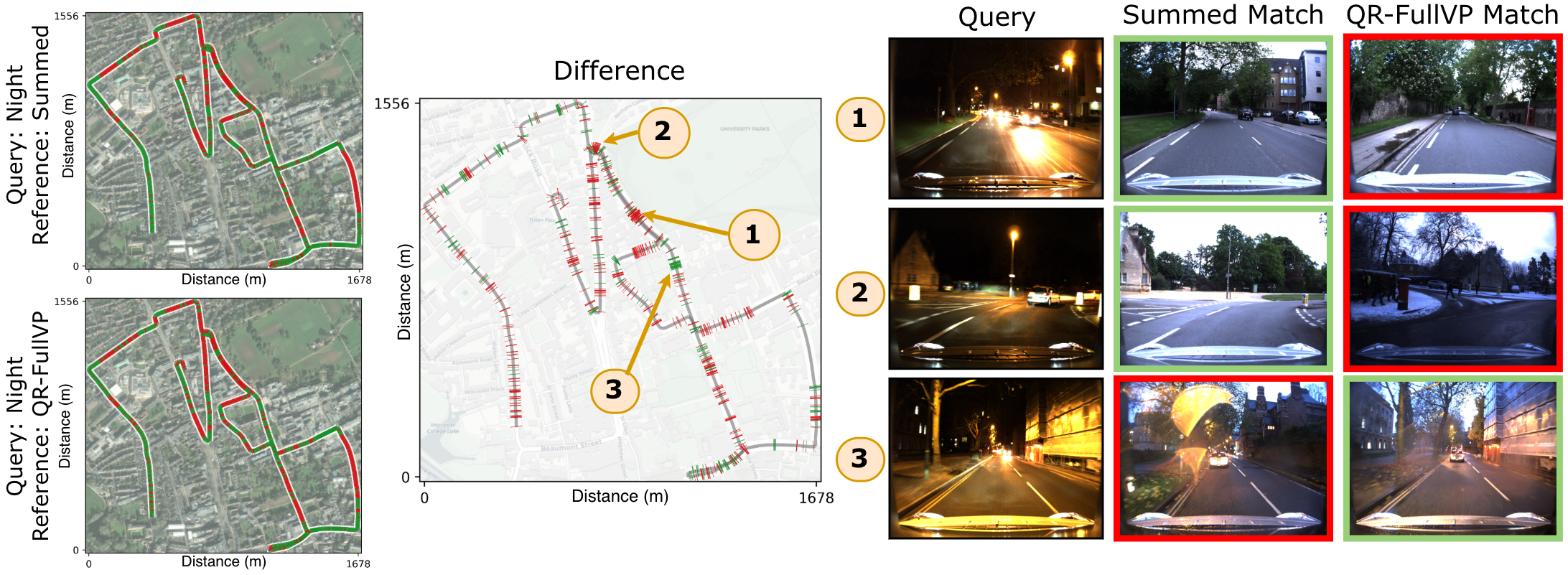}
    \caption{
        \textbf{Retrieval comparison on Oxford RobotCar (night query) using Eigenplaces.} Left shows retrieval outcomes for reference descriptor summing and QR-FullVP, with green indicating correct and red incorrect top-1 matches. The difference map (center) highlights where QR-FullVP succeeds over summing (green), the reverse (red), or both agree (grey). Although outperforming pooling, summing produces a more robust aggregate under night conditions, likely because this severe appearance shift is not as well represented by the span of the reference basis, and benefiting from reinforcement of low-variance structure. However, summing relies on viewpoint consistency across references, with performance degrading when viewpoints differ within a place's collection or where outlier effects such as light refraction are present, as in the third example. (Retrieved images for Summed \& QR-FullVP show a randomly sampled image from the matched place's reference collection).}
    \label{fig:map3}
    \vspace*{-1em}
\end{figure*}

Despite this, several challenging cases such as Nordland-Winter and ORC-Night queries benefited more from heuristic aggregation like score averaging or descriptor summing. We suspect this occurs when queries exhibit extreme appearance shifts that are not as well approximated by the reference basis. If viewpoint remains relatively fixed, aggregation-based methods may provide additional robustness by smoothing appearance-specific variations and reinforcing any persistent geometric structure. Projection residuals across datasets support this (\fig{fig:residuals}), where although correct matches consistently yield lower residuals, these challenging traverses exhibit notably larger residuals overall. As shown in \fig{fig:map3}, summing descriptors occasionally outperforms QR-FullVP under night conditions, consistent with queries lying further from the reference basis despite both still outperforming pooling. However, in our qualitative analysis we found that aggregation methods break down on Oxford RobotCar when place references are captured from opposing directions, or under extreme lighting effects such as refraction. Our method achieves strong overall performance, and, as shown in the following results, effectively handles both viewpoint variation and scenes with unconstrained references, providing a general and robust solution for multi-reference VPR.

\subsection{Multi-Viewpoint Datasets}

\begin{figure}[t]
    \centering
    \includegraphics[width=\columnwidth]{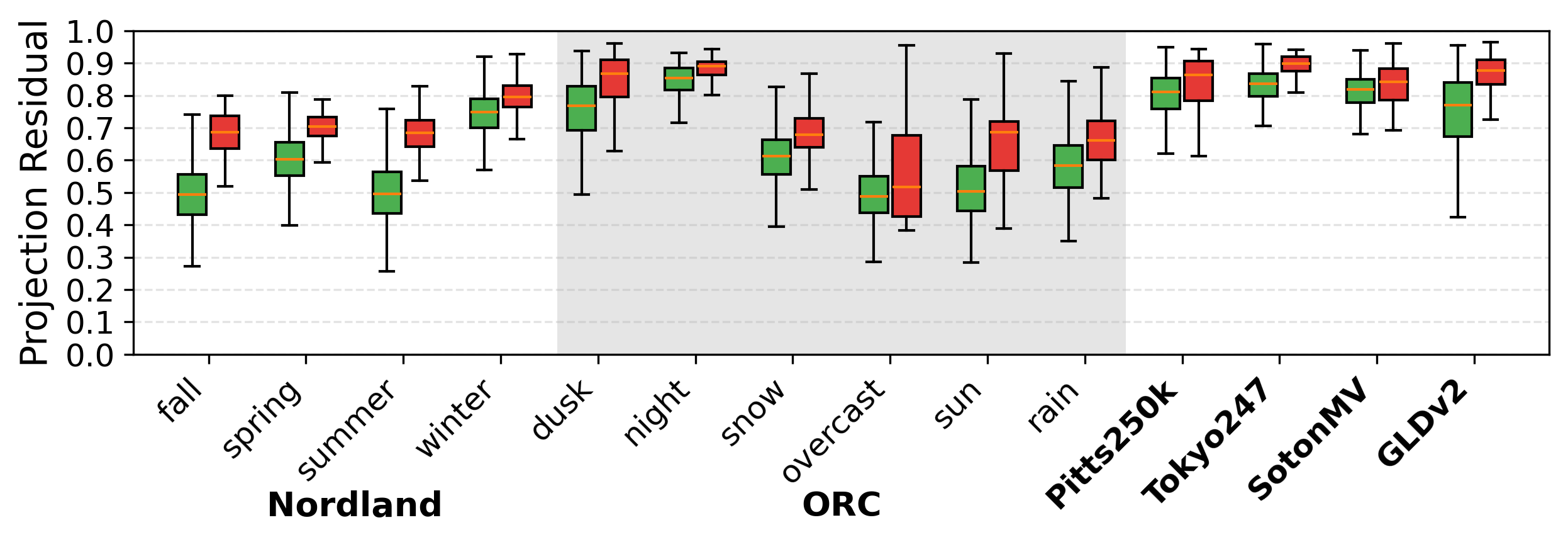}
    \caption{
        \textbf{QR-FullVP residuals using Eigenplaces for correct (green) and incorrect (red) top-1 matches across all datasets.} Correct matches show lower residuals across all conditions. Appearance-challenging datasets (ORC Night, Nordland Winter), show larger overall residuals, reflecting a larger domain gap between query and reference basis. Multi-viewpoint datasets exhibit high residuals (from non-overlapping viewpoints), yet still improve retrieval over pooling, suggesting the subspace is sufficiently discriminative.}
    \label{fig:residuals}
    \vspace{-1em}
\end{figure}

\begin{figure*}[!ht]
    \centering
    \includegraphics[width=0.91\textwidth]{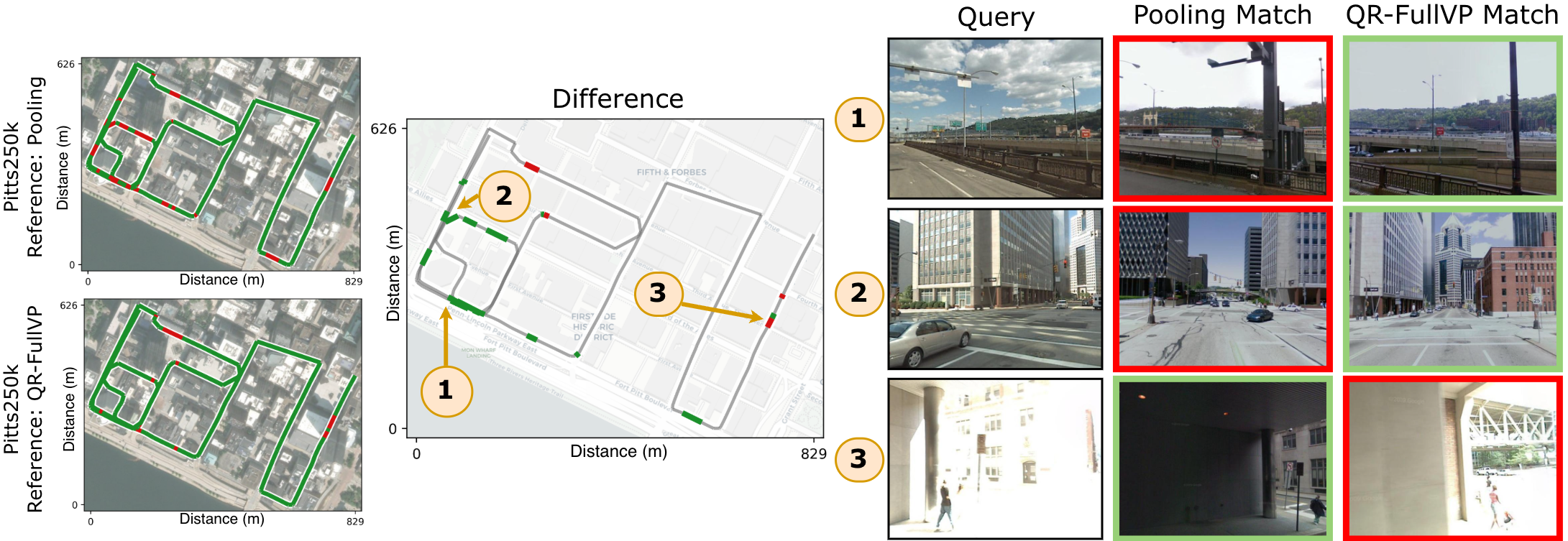}
    \caption{
        \textbf{Retrieval comparison on Pitts250k-Test using Eigenplaces.} Left shows retrieval outcomes for reference pooling and QR-FullVP, with green indicating correct and red incorrect top-1 matches. The difference map (center) highlights where QR-FullVP succeeds over pooling (green), the reverse (red), or both agree (grey). QR-FullVP outperforms pooling, using complementary viewpoint information, and qualitative examples suggest that QR-FullVP improves disambiguation, retrieving places closer to the true location. (Retrieved images for QR-FullVP show a randomly sampled image from the matched place's reference collection).}
    \label{fig:map4}
    \vspace*{-1em}
\end{figure*}

Tables \ref{tab:multiview-soton}, \ref{tab:multiview-soton2}, \ref{tab:multiview-pitts}, \ref{tab:multiview-pitts2}, \ref{tab:multiview-tokyo}, and \ref{tab:multiview-tokyo2} compare the performance of our QR-based approach on multi-viewpoint datasets, namely SotonMV, Pitts250k-Test and Tokyo24/7. We reiterate that for SotonMV and Pitts250k-Test, we filter query images at intermediate headings between the reference orientations, isolating each approaches' robustness to poor and distributed spatial overlap. For these two datasets, our approach achieves up to $\approx$10\% improvement in Recall@1 across all VPR methods, consistently improving performance over pooling unlike aggregation approaches, which as expected, are intolerant to viewpoint shift as the unique spatial contexts are collapsed in the aggregated representation. The performance gains of the lighter CNN-based models are generally larger than with the heavier transformer models, reflecting recent work with large-scale training and robust backbones for inherently improved invariance. LogSumExp re-ranking exhibits marginal performance gains, highlighting the value of such a two-stage weighting approach. As shown in the traverse in \fig{fig:map4}, QR-FullVP greatly outperforms pooling, with examples suggesting improved metric precision in retrievals. From \fig{fig:residuals}, we find that the projection residuals of our approach for multi-view evaluation are higher, which is expected as each place's subspace accounts for a range of camera yaw angles. Nevertheless, QR-FullVP appears to preserve each place's intrinsic viewpoint variation and discriminability, enabling query descriptors to be matched via projection onto this span. Results with Tokyo24/7 are more nuanced. For Tokyo24/7, reference place images are also captured at 60° intervals in the day, however, while queries have variable degrees of viewpoint change with translations, they emphasize appearance changes (day, sunset, night) and severe structural changes in a dense urban environment. We can see that with QR-FullVP, Recall@1 improves for all lighter CNN models, particularly Eigenplaces which emphasizes viewpoint invariance. However, for heavier transformer models, SALAD and BoQ show marginal gains while MegaLoc and CricaVPR benefit more with conventional pooling. We suspect the reference collection of Tokyo24/7 limits our approach here. The viewpoint-only references do not account for the severe query appearance variation, and for models like CricaVPR, these subspaces may suppress query descriptor components useful for matching that direct cosine comparison preserves. Despite modest viewpoint shifts in Tokyo24/7, we expect further benefit if query appearance variations are included in the reference collections.

\subsection{Unconstrained References}

\begin{figure*}[ht]
    \centering
    \includegraphics[width=0.87\textwidth]{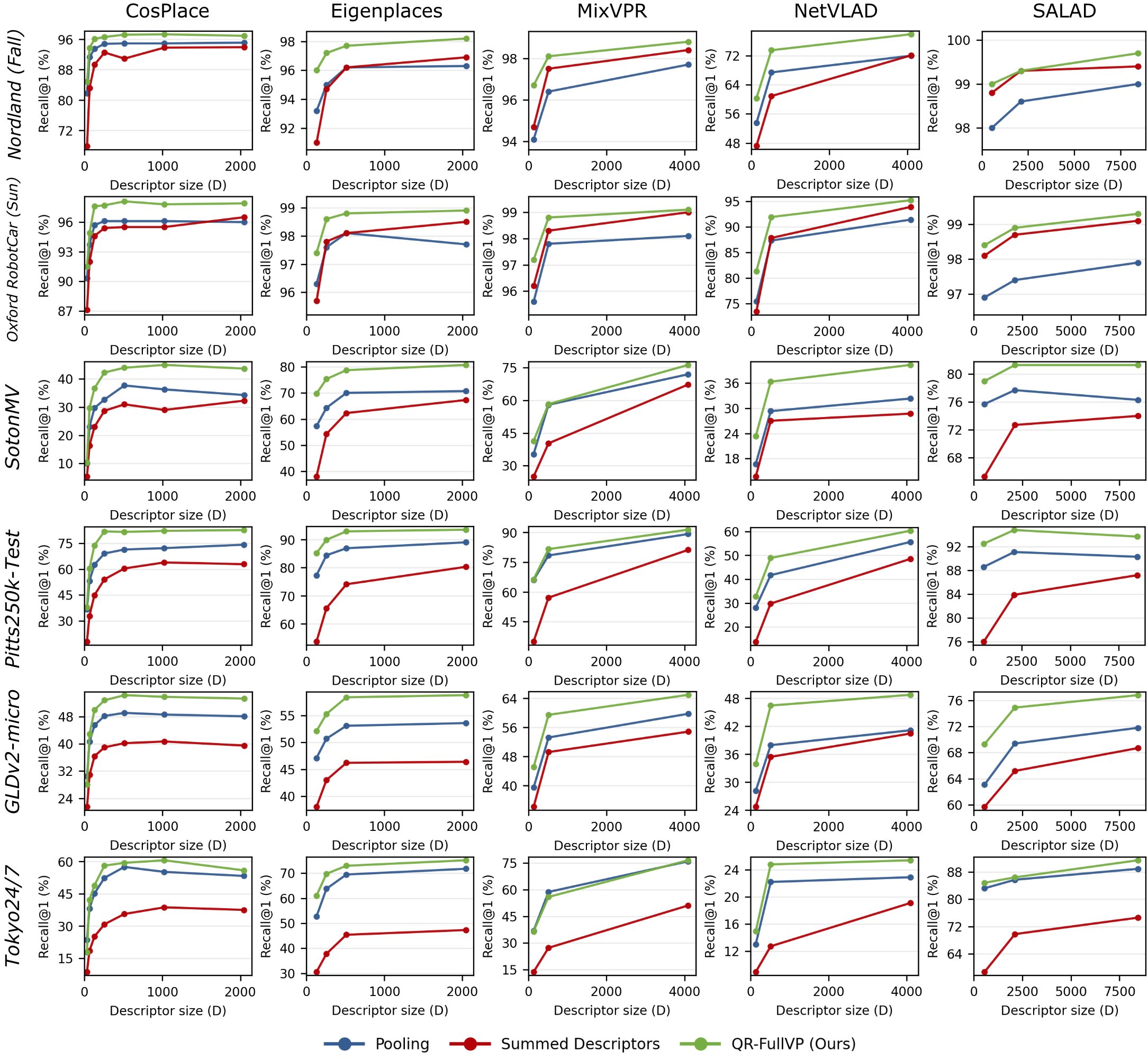}
    \caption{
        Recall@1 versus descriptor size. Our QR-based approach using smaller descriptor sizes often outperforms other multi-reference approaches at maximum size, offering optimization with map memory. \textbf{Note:} for each plot the y-axis is scaled respectively for clear visualization.
    }
    \label{fig:descsize}
    \vspace*{-1em}
\end{figure*}

From the results on the Google Landmarks dataset (GLDv2) in \tabs{tab:gldv2}{tab:gldv2_2}, our approach achieves a consistent improvement of $\approx$5\% Recall@1 over the next best performing multi-reference approach, across all models. Unlike the multi-reference datasets where appearance or viewpoint change is regularly structured, GLDv2 consists of reference sets with heterogeneous viewpoints, translations, and image properties, closely resembling real-world localization scenarios where controlled acquisition is not possible.

While the other multi-reference approaches like aggregation provide modest gains on appearance-change datasets, they under perform even relative to pooling in this unconstrained reference setting, highlighting their dependence on viewpoint alignment and consistency. In contrast, our method demonstrates stronger generalization and reliability, effectively utilizing all available reference information as a unified subspace, regardless of capture conditions and without requiring structured data acquisitions.

\subsection{Sequence-Based VPR}
\tab{tab:seqvlad} evaluates SeqVLAD on Nordland, Oxford-RobotCar and Pitts250k-Test, which we use due to their size and sequence-based structure. Our approach is implemented in the same manner. For Nordland and Oxford-RobotCar, each traverse is grouped into subsequences of 5 consecutive frames, each producing a single 24576D descriptor. For Pitts250k-Test, subsequences are formed from 5 consecutive frames at each respective viewpoint angle, producing six 24576D descriptors per place. In both cases, the resulting descriptors are used directly as columns of the place matrix for QR-FullVP, naturally incorporating both temporal information from the model and multi-viewpoint information into the subspace. QR-FullVP consistently outperforms all single and most multi-reference baselines across all three datasets, demonstrating that the subspace approach generalizes effectively with the sequence-based descriptors. Like with other methods however, Nordland Winter and Oxford-RobotCar Night are particular challenging cases, with descriptor summation providing a marginal gain.

\subsection{Dimensionality Reduction}
\begin{figure*}[!t]
  \centering

  \includegraphics[width=0.75\textwidth]{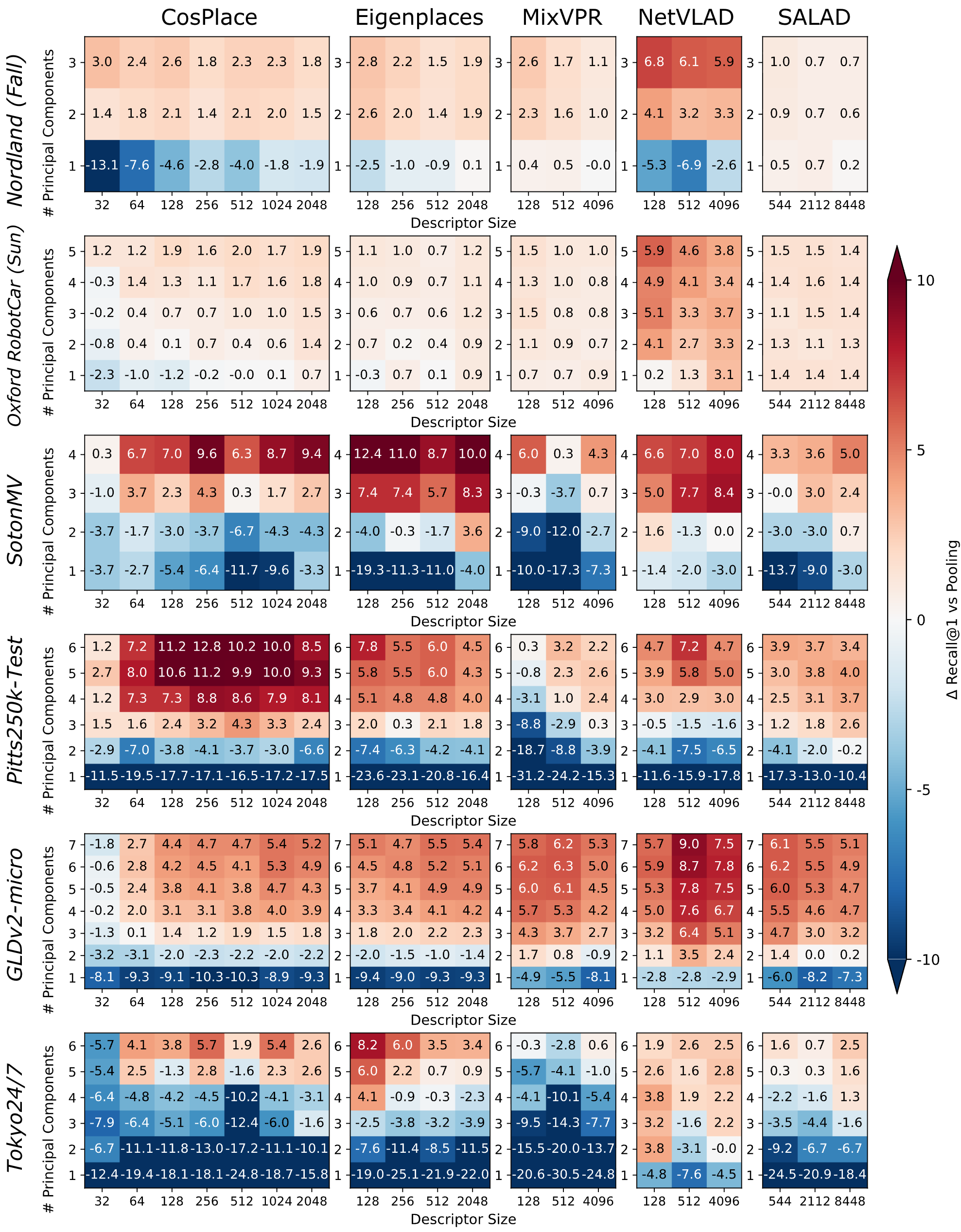}
  \caption{\textbf{Change in Recall@1} against pooling (\textit{with the same descriptor size}) using SVD decomposition instead of QR. At full rank (top rows), the performance is equivalent to QR-FullVP.}
  \vspace*{-1em}
  \label{fig:svd}
\end{figure*}

Aggregation-based methods like averaging, while fragile to viewpoint, improve computational efficiency by reducing multiple place descriptors to a single descriptor representation, reducing map memory usage and computation. We investigate two strategies to reduce the computational overhead of our framework. First, we analyze the effect of descriptor size reduction and assess performance sensitivity to dimensionality. For this, we use models with pre-trained weights available across multiple embedding sizes. Second, we apply SVD for rank reduction, keeping the dominant direction of variations across each place's reference descriptors and discarding low-variance components for each subspace, reducing computation and memory.

\textbf{Descriptor Size Analysis.} Generally, increasing descriptor dimensionality improves performance for all methods. With feature extractors like CosPlace, performance levels off and occasionally even diminishes with larger descriptor size (as expected from \cite{9880209}) across all matching approaches, reflecting properties of the underlying embeddings and its suitability to the dataset. Our decomposition framework remains robust and effective under aggressive descriptor compression, often surpassing other multi-reference approaches at full descriptor dimensionality (see \fig{fig:descsize}). For example, Eigenplaces with descriptor length reduced by a factor of four drops by only 1\% Recall@1, while CosPlace with 16-fold descriptor compression still outperforms uncompressed baselines. NetVLAD, MixVPR and SALAD are capable of maintaining strong performance with 8-fold compression. Our approach can maintain its performance advantage even under substantial descriptor size reduction, enabling significantly more efficient deployment.
    
\textbf{SVD Rank Reduction.} \fig{fig:svd} compares performance against pooling when using SVD in place of QR for place decomposition, which enables rank truncation. The principal components of each place's subspace are linear combinations of its reference descriptors, encoding the directions of greatest variation across appearance and viewpoint. Truncation therefore discards directions that contribute least to explaining differences among the references (though not necessarily least discriminative). We measure the sensitivity of retrieval performance to rank truncation. With all principal components retained, performance is identical to QR, as both span the same subspace. As components are discarded, performance generally degrades, with more pronounced degradation on the multi-view datasets where the references capture non-overlapping viewpoints that each contribute distinct directions in the subspace that a query may depend on. Nevertheless, for certain configurations, removing one to three components yields near-identical performance gain with reduced storage and matching cost. While retaining all components is typically advantageous, discarding these lower-variance components can have a small impact on representation quality with practical per-place compression.

\begin{figure}[!t]
    \centering
    \includegraphics[width=\linewidth]{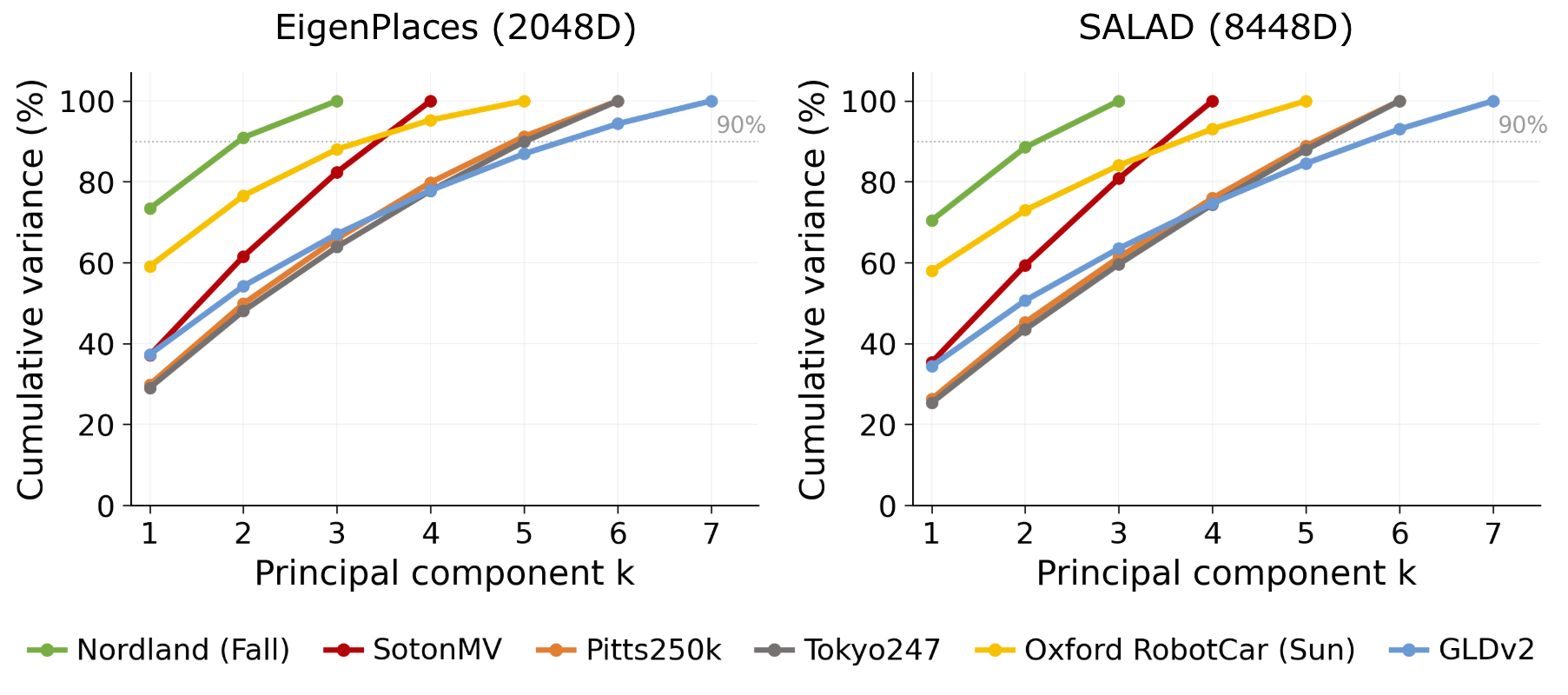}
    \vspace{-2em}
    \caption{%
        Average
        cumulative variance of SVD-derived principal components for reference representations across the datasets. Nordland and Oxford RobotCar exclude Fall and Sun conditions, respectively.
    }
    \label{fig:spectrum}
    \vspace{-1em}
\end{figure}

From \fig{fig:spectrum}, we observe that appearance-change datasets exhibit a larger dominant component of variation, since all reference images share the same viewpoint. For example, Nordland's first component captures around 70\% of the variance. In contrast, viewpoint-change datasets show a flatter distribution, as expected since the reference viewpoints are distinct and non-overlapping, and so are more sensitive to truncation. Consistent with the results, keeping components that capture 80-90\% of the cumulative variance can preserve the observed performance comparable to the full subspace, where the reference information remains well-approximated by the truncated subspace.

\subsection{Computational Overhead}

Our approach introduces no additional training cost and remains independent of the underlying VPR feature extractor (see \tab{tab:overhead}), therefore, extraction time is unchanged. For mapping with $D$-dimensional descriptors and $N$ reference images per place, QR decomposition has complexity $\mathcal{O}(DN^2)$, and the full SVD $\mathcal{O}(DN^2 + N^3)$. For $M$ total places, these are $\mathcal{O}(MDN^2)$ and $\mathcal{O}(MDN^2 + MN^3)$, respectively. Since all approaches ultimately operate over the same set of reference images, matching via cosine distance (pooling) or subspace projection (QR and SVD approach) has identical asymptotic complexity dominated by the matrix-vector product $\mathcal{O}(MDN)$, followed by a lower-order aggregation like max or norm. 

\begin{figure}[!t]
    \centering
\includegraphics[width=0.95\linewidth]{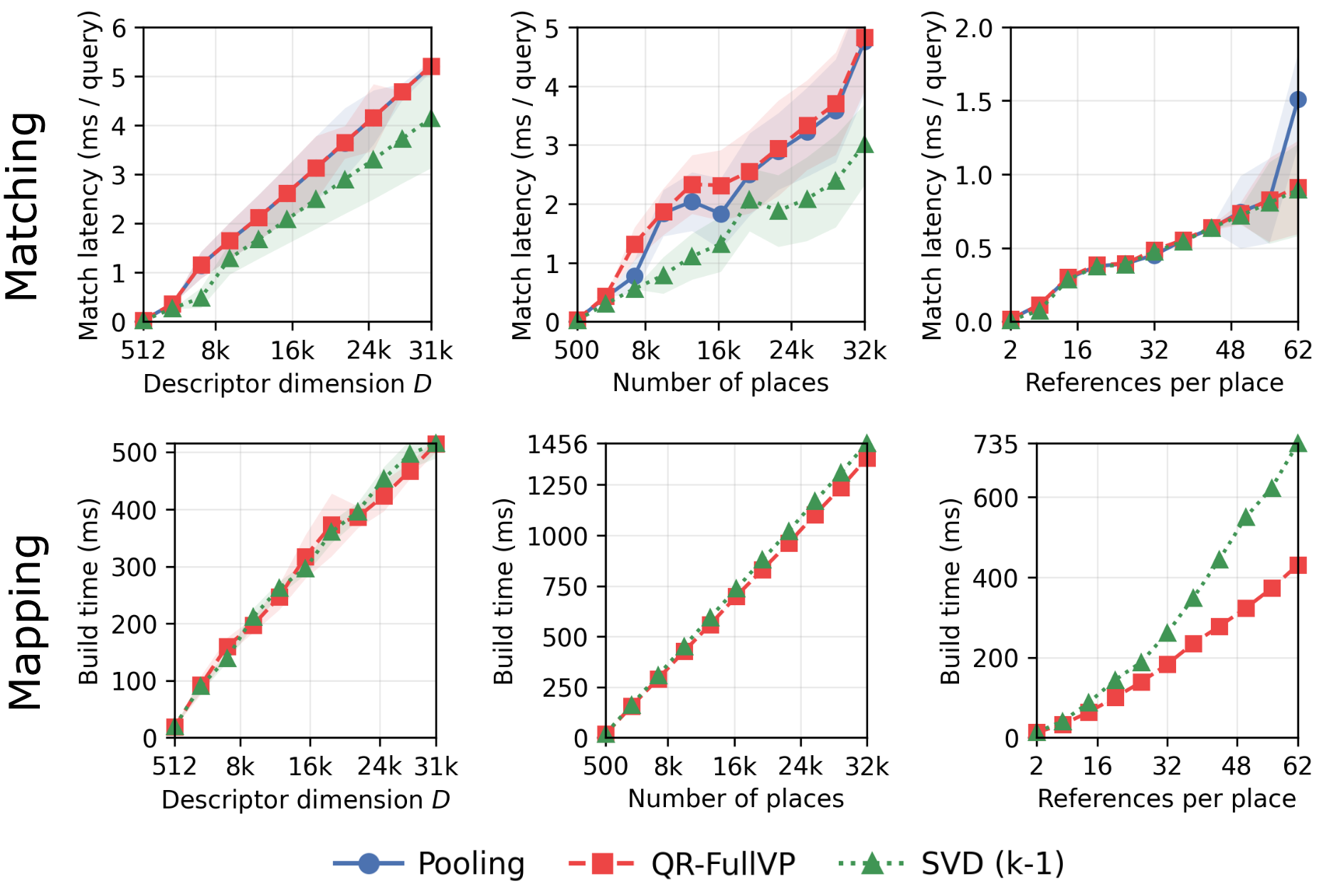}
    \caption{Match and map construction latency for pooling, QR-FullVP, and SVD (k-1) evaluated on an Apple M4 CPU using SciPy linear algebra with synthetic descriptors. Queries are processed sequentially (batch size 1) with single-threaded BLAS, and all approaches use identical matrix-vector multiplication for matching, differing only in the post-multiplication aggregation. Map construction is performed per place via QR or truncated SVD. Results are reported as median latency over 300 queries across sweeps in descriptor dimension, number of places, and references per place, which, when non-swept are fixed to 512, 500, and 4 respectively.}
    \label{fig:overheadCPU}
    \vspace{-1em}
\end{figure}

\fig{fig:overheadCPU} confirms this behavior, where we measure the latency of each approach using synthetic descriptors on CPU, implementing all approaches with the same matrix multiplication followed by their respective aggregation with single-threaded BLAS. Matching latency exhibits a consistent linear trend with increasing problem size, with truncated SVD slightly reducing match time due to the reduced subspace dimension. For map construction, QR and SVD show comparable scaling with descriptor dimension $D$ and number of places $M$, reflecting their dominance with tall matrices ($D>>N$). As the number of references per place increases, SVD starts to diverge as the power terms begin to dominate, although practically this is limited as our places are not represented with as many images. Despite maintaining efficiency, we note that the proposed subspace-based matching is not directly compatible with approximate nearest neighbor (ANN) indexing typically used in very large-scale systems. A simple workaround could involve an initial coarse retrieval stage with ANN prior to subspace-based re-ranking.

\tab{tab:overhead} shows performance overhead with SotonMV and Pitts250k-Test. We highlight that larger discrepancies between approaches arise from implementation decisions. Unlike \fig{fig:overheadCPU}, pooling is executed as a direct matrix-vector multiplication, while for the subspace methods we use einsum-based projections, resulting in less efficient kernel and BLAS behavior. Similarly mapping uses full decompositions per place with additional overhead from GPU execution and synchronization. Nevertheless, matching remains efficient, relying only on projection magnitudes, and at lower descriptor dimensionality it is faster, more memory efficient and superior to pooling with larger descriptor sizes. Map construction, pre-computed, scales reasonably with the number of reference images, supporting low-latency updates. Finally, truncated SVD can further reduce map memory and matching cost, though potentially affecting retrieval performance, as observed earlier.

\newcommand{\vhdr}[1]{
  \makecell{\rule{0pt}{2.6ex}#1\rule[-1.3ex]{0pt}{0pt}}%
}

\begin{table*}[!ht]
\centering
\footnotesize
\setlength{\tabcolsep}{3.5pt}    
\renewcommand{\arraystretch}{1.05} 

\caption{Computational overhead of conventional VPR vs QR and SVD-based approach using an Nvidia A100 GPU with a batch size of 1. QR uses all ($k$) references for the subspace while SVD truncates to $k-1$. Note that matching and map construction have asymptotic complexity across approaches, however, measured differences reflect our implementation choices and kernel behavior, with pooling executed as a matrix–vector multiplication, subspace methods using einsum-based projections with reductions, map construction performed via full per-place decompositions and explicit synchronizations, affecting the practical runtime.}
\label{tab:overhead}
\begin{tabular}{|l|c|l|c|cccc|cccc|}
\hline
\multirow{3}{*}{Method} & \multirow{3}{*}{Dim} & \multirow{3}{*}{Approach} &
\multirow{3}{*}{\parbox{1.25cm}{\centering Extraction\\ Time\\(ms/query)}} &
\multicolumn{4}{c|}{\textbf{SotonMV} (300 Query, 63872 Ref. Images)} &
\multicolumn{4}{c|}{\textbf{Pitts250k Test} (1000 Query, 20988 Ref. Images)}\\
\cline{5-12}
 &  &  &  &
\vhdr{Match Time\\(ms/query)} &
\vhdr{Total Map\\Time (s)} &
\vhdr{Map Mem.\\(GB)} &
\vhdr{R@1} &
\vhdr{Match Time\\(ms/query)} &
\vhdr{Total Map\\Time (s)} &
\vhdr{Map Mem.\\(GB)} &
\vhdr{R@1}  \\
\hline\hline

CosPlace & 2048 & Pooling & 5.90 & 0.43 & - & 0.523 & 34.2  & 0.15 & - & 0.172 & 74.2  \\
CosPlace & 2048 & QR-FullVP & 5.57 & 1.41 & 3.22 & 0.523 & 43.7 & 0.48 & 0.88 & 0.172 & 82.7\\
CosPlace & 2048 & SVD (k-1) & 4.78 & 1.20 & 9.25 & 0.392 & 36.9 & 0.39 & 2.50 & 0.143 & 83.5\\
\hline
CosPlace & 256 & Pooling & 6.21 & 0.12 & - & 0.0654 & 32.7 & 0.05 & - & 0.0215 & 69.1  \\
CosPlace &  256 & QR-FullVP & 5.61 & 0.26 & 1.66 & 0.0654 & 42.3 & 0.11 & 0.45 & 0.0215 & 81.9 \\
CosPlace &  256 & SVD (k-1) & 4.98 & 0.23 & 6.54 & 0.0491 & 37.0 & 0.09 & 1.83 & 0.0179 & 80.3 \\

\hline
\end{tabular}
\vspace*{-1em}
\end{table*}

\subsection{Orientation Recovery}

\begin{figure}[!t]
    \centering
    \includegraphics[width=\linewidth]{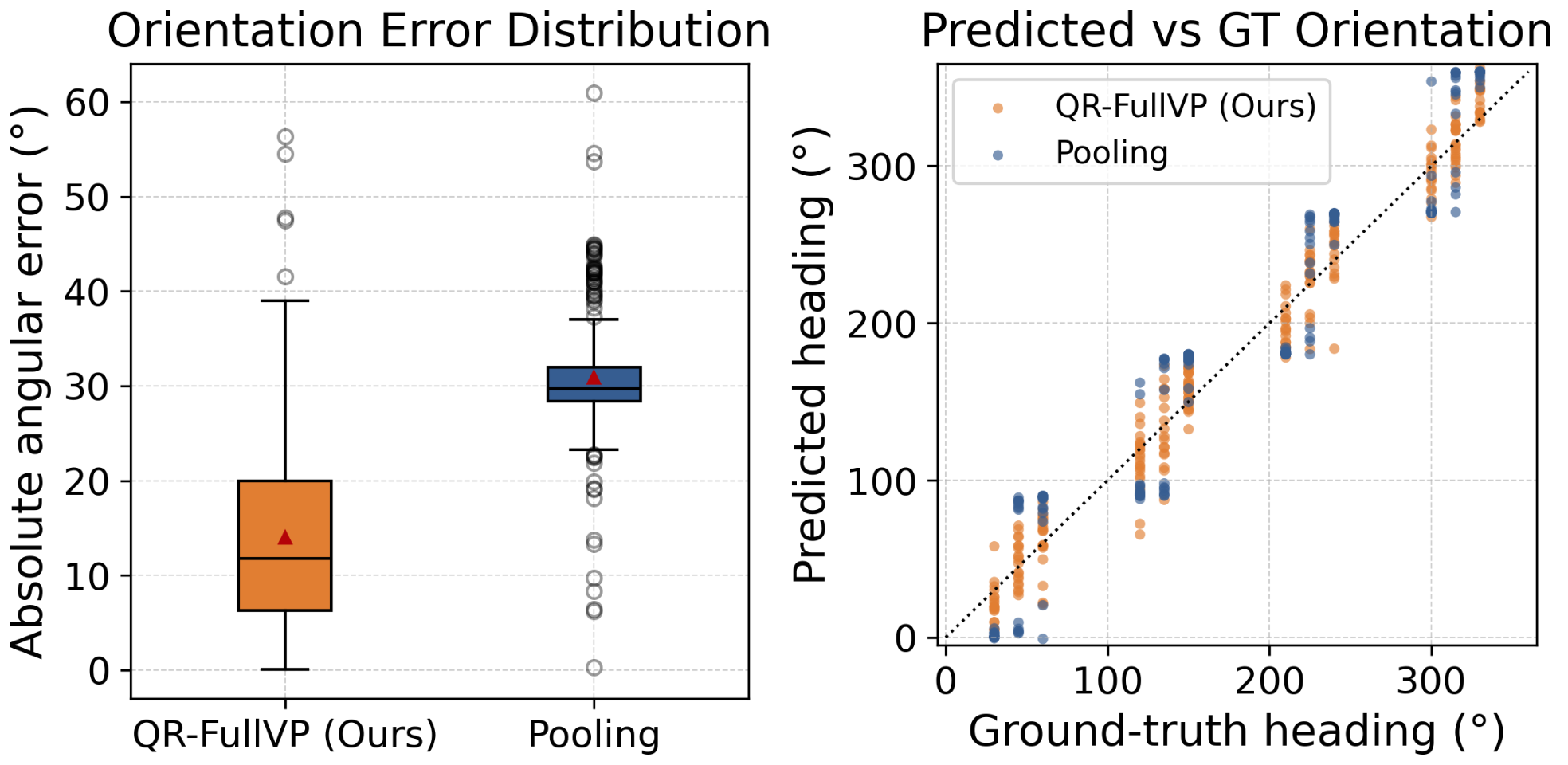}
    \caption{%
        Evaluation of orientation estimates with SotonMV.
        (\textit{Left}) Distribution of absolute angular errors. 
        (\textit{Right}) Predicted vs ground-truth headings.
    }
    \label{fig:orientation_eval}
    \vspace{-1em}
\end{figure}

We investigate the estimation of relative camera yaw solely from appearance, without geometric computation, and assuming the SotonMV map setup, where each place is represented by images taken at regular intervals of yaw. The
$\textbf{R}$ matrix from QR decomposition linearly relates the projected query coefficients to the original heading basis, giving coefficients that reflect the relative contributions of the four headings. We interpret these coefficients as weights and use circular interpolation to produce a continuous orientation estimate from the discrete headings. We compare this with pooling, where we apply softmax to the per-heading similarity scores and use these weights for circular averaging. Because queries are sampled within $5m$ of their ground truth reference locations, translation induces a systematic bias between estimated and true orientations, which can be geometrically upper bounded as:
\begin{equation}
\label{eq:orientation_bias_bound}
|\Delta\theta|_{\max} \;=\;
\arctan\!\left(
    \frac{T}{\sqrt{D^{2} - T^{2}}}
\right),
\qquad (T < D),
\end{equation}

where $T$ is the maximum query translation magnitude and $D$ the effective depth of view. With a maximum $T=5m$ and approximating a minimum $D=10m$ for street-view imagery, we approximate the bias bound at 30°, and so consider correct estimations to be within a ±30° threshold. \fig{fig:orientation_eval} shows the estimated and ground truth headings, and orientation error distribution using Eigenplaces with 2048-dimensional descriptors restricted to correctly matched queries. With QR, we achieve an estimation accuracy of 95.0\% (230/242 queries within the threshold) and a mean absolute error of 13.8°, while with pooling, we get an accuracy of 67.9\% (144/212) and a mean error of 30.9°. This formulation requires that reference headings are known, captured from a fixed location per place and that queries are localized sufficiently close to their matched reference. Under these conditions, relative orientation can be recovered from arbitrarily spaced headings via interpolation, with denser angular sampling improving coverage. The results indicate that orientation is reasonably recoverable from appearance alone, and that orientation is more effectively recovered with the QR-based approach, supporting the assumption that descriptor variation with viewpoint can be well-approximated by a locally linear subspace.

\newcolumntype{C}{>{\centering\arraybackslash}c}

\renewcommand{\rot}[1]{%
  \raisebox{0pt}[2.0em][1.5em]{\adjustbox{angle=90,valign=c}{#1}}%
}

\begin{table*}[!ht]
\centering
\scriptsize
\caption{Recall@1 with Nordland retrieval using pairwise reference-season combinations for place representations.}
\label{tab:nordlandpairs}

\setlength{\tabcolsep}{0.7pt}
\renewcommand{\arraystretch}{1.25}

\resizebox{\textwidth}{!}{%
\begin{tabular}{|>{\raggedright\arraybackslash}p{2.0cm}|
  C C C !{\vrule width 0.4pt} C C C !{\vrule width 0.4pt} C C C !{\vrule width 0.4pt} C C C |
  C C C !{\vrule width 0.4pt} C C C !{\vrule width 0.4pt} C C C !{\vrule width 0.4pt} C C C |
  C C C !{\vrule width 0.4pt} C C C !{\vrule width 0.4pt} C C C !{\vrule width 0.4pt} C C C |}
\hline
\textbf{Methods $\rightarrow$} &
\multicolumn{12}{c|}{\textbf{EigenPlaces (2048D)}} &
\multicolumn{12}{c|}{\textbf{CosPlace (2048D)}} &
\multicolumn{12}{c|}{\textbf{MixVPR (4096D)}} \\
\cline{2-37}
\textbf{Queries $\rightarrow$} &
\multicolumn{3}{c|}{\textbf{Fall}} &
\multicolumn{3}{c|}{\textbf{Spring}} &
\multicolumn{3}{c|}{\textbf{Summer}} &
\multicolumn{3}{c|}{\textbf{Winter}} &
\multicolumn{3}{c|}{\textbf{Fall}} &
\multicolumn{3}{c|}{\textbf{Spring}} &
\multicolumn{3}{c|}{\textbf{Summer}} &
\multicolumn{3}{c|}{\textbf{Winter}} &
\multicolumn{3}{c|}{\textbf{Fall}} &
\multicolumn{3}{c|}{\textbf{Spring}} &
\multicolumn{3}{c|}{\textbf{Summer}} &
\multicolumn{3}{c|}{\textbf{Winter}} \\
\cline{2-37}
\textbf{References $\rightarrow$} &
\rot{Sp+Su} & \rot{Sp+Wi} & \rot{Su+Wi} &
\rot{Fa+Su} & \rot{Fa+Wi} & \rot{Su+Wi} &
\rot{Fa+Sp} & \rot{Fa+Wi} & \rot{Sp+Wi} &
\rot{Fa+Sp} & \rot{Fa+Su} & \rot{Sp+Su} &
\rot{Sp+Su} & \rot{Sp+Wi} & \rot{Su+Wi} &
\rot{Fa+Su} & \rot{Fa+Wi} & \rot{Su+Wi} &
\rot{Fa+Sp} & \rot{Fa+Wi} & \rot{Sp+Wi} &
\rot{Fa+Sp} & \rot{Fa+Su} & \rot{Sp+Su} &
\rot{Sp+Su} & \rot{Sp+Wi} & \rot{Su+Wi} &
\rot{Fa+Su} & \rot{Fa+Wi} & \rot{Su+Wi} &
\rot{Fa+Sp} & \rot{Fa+Wi} & \rot{Sp+Wi} &
\rot{Fa+Sp} & \rot{Fa+Su} & \rot{Sp+Su}\\
\hline\hline
\textit{Pooling} &
96.4 & 80.7 & 94.4 & 91.4 & 91.4 & 87.2 & 93.6 & 93.3 & 73.7 & 69.1 & 67.3 & 69.5 &
95.1 & 76.8 & 93.0 & 88.4 & 89.0 & 85.0 & 92.1 & 91.2 & 68.7 & 69.8 & 61.7 & 69.8 &
97.7 & 88.1 & 95.8 & 97.0 & 97.6 & 95.0 & 96.4 & 95.9 & 83.3 & 90.7 & 85.4 & 91.4 \\

\hline
\textit{dMat. Avg.} &
95.9 & 76.2 & 90.7 & 90.0 & 91.8 & 89.1 & 91.1 & 87.5 & 68.7 & \textbf{76.6} & 66.9 & \textbf{75.8} &
92.7 & 69.7 & 83.6 & 87.3 & 86.2 & 83.6 & 87.2 & 80.2 & 62.7 & \textbf{73.4} & 61.1 & 71.6 &
97.6 & 90.1 & 96.6 & 96.2 & 98.1 & 96.6 & 95.3 & 94.8 & 85.2 & \textbf{93.2} & 84.2 & 92.9 \\
\textit{LogSumExp} & 97.0 & 81.7 & 95.1 & 90.9 & 94.1 & 90.5 & 93.5 & 93.5 & 74.8 & 74.8 & 67.6 & 74.3 & 95.2 & 77.6 & 93.5 & 88.7 & 90.4 & 87.0 & 91.6 & 91.8 & 69.3 & 73.0 & \textbf{62.3} & 73.1 & 98.1 & 89.9 & 96.6 & 96.7 & 98.3 & 96.6 & 96.2 & 95.9 & 84.9 & 93.3 & 85.2 & \textbf{93.4}\\
\textit{Sum Descriptors} &
97.8 & 80.4 & 95.4 & 91.5 & 95.8 & 93.9 & 93.3 & 92.3 & 73.2 & 75.2 & 67.6 & 75.0 &
96.3 & 75.3 & 91.5 & 88.7 & 92.6 & 90.7 & 91.4 & 87.7 & 68.2 & 72.4 & 61.5 & \textbf{71.8} &
98.6 & \textbf{90.9} & \textbf{97.7} & 97.3 & 98.5 & 97.7 & 96.5 & 96.1 & \textbf{86.3} & 92.7 & 86.4 & 92.9 \\
\hline
\textit{\textbf{QR-FullVP (Ours)}} &
\textbf{98.0} & \textbf{82.3} & \textbf{96.4} & \textbf{92.3} & \textbf{96.0} & \textbf{94.1} & \textbf{94.6} & \textbf{93.9} & \textbf{75.7} & 71.9 & \textbf{68.0} & 71.6 &
\textbf{96.6} & \textbf{78.8} & \textbf{94.9} & \textbf{89.4} & \textbf{93.7} & \textbf{91.9} & \textbf{93.1} & \textbf{92.0} & \textbf{70.6} & 71.2 & 62.1 & 70.9 &
\textbf{98.7} & 90.2 & 97.4 & \textbf{97.4} & \textbf{98.7} & \textbf{97.8} & \textbf{96.7} & \textbf{96.3} & 85.5 & 92.4 & \textbf{86.5} & 93.2\\
\hline
\end{tabular}%
} 
\end{table*}

\renewcommand{\rot}[1]{%
  \raisebox{0pt}[2.0em][1.5em]{\adjustbox{angle=90,valign=c}{#1}}%
}
\begin{table*}[!ht]
\centering
\fontsize{5.25pt}{6.3pt}\selectfont
\caption{Recall@1 with Oxford Robot Car retrieval using pairwise reference-weather combinations for place representations.}
\label{tab:oxfordpairs}
\setlength{\tabcolsep}{1pt}
\renewcommand{\arraystretch}{1.25}
\begin{tabular}{|>{\raggedright\arraybackslash}p{2.0cm}|
  C C C C C C C C C C |
  C C C C C C C C C C |
  C C C C C C C C C C |
  C C C C C C C C C C |}
\hline
\textbf{Methods $\rightarrow$} &
\multicolumn{10}{c|}{\textbf{MegaLoc (8448D)}} &
\multicolumn{10}{c|}{\textbf{CricaVPR (10752D)}} &
\multicolumn{10}{c|}{\textbf{BoQ (12288D)}} &
\multicolumn{10}{c|}{\textbf{SALAD (8448D)}} \\
\cline{2-41}
\textbf{Queries $\rightarrow$} &
\multicolumn{10}{c|}{\textbf{Sun}} &
\multicolumn{10}{c|}{\textbf{Overcast}} &
\multicolumn{10}{c|}{\textbf{Night}} &
\multicolumn{10}{c|}{\textbf{Snow}} \\
\cline{2-41}
\textbf{References $\rightarrow$} &
\rot{Du+Ra} & \rot{Du+Ov} & \rot{Du+Ni} & \rot{Du+Sn} &
\rot{Ra+Ov} & \rot{Ra+Ni} & \rot{Ra+Sn} &
\rot{Ov+Ni} & \rot{Ov+Sn} &
\rot{Ni+Sn} &
\rot{Su+Du} & \rot{Su+Ra} & \rot{Su+Ni} & \rot{Su+Sn} &
\rot{Du+Ra} & \rot{Du+Ni} & \rot{Du+Sn} &
\rot{Ra+Ni} & \rot{Ra+Sn} &
\rot{Ni+Sn} &
\rot{Su+Du} & \rot{Su+Ra} & \rot{Su+Ov} & \rot{Su+Sn} &
\rot{Du+Ra} & \rot{Du+Ov} & \rot{Du+Sn} &
\rot{Ra+Ov} & \rot{Ra+Sn} &
\rot{Ov+Sn} &
\rot{Su+Du} & \rot{Su+Ra} & \rot{Su+Ov} & \rot{Su+Ni} &
\rot{Du+Ra} & \rot{Du+Ov} & \rot{Du+Ni} &
\rot{Ra+Ov} & \rot{Ra+Ni} &
\rot{Ov+Ni} \\
\hline\hline
\textit{Pooling} & 93.1 & 95.8 & 73.2 & 88.8 & 97.0 & 93.6 & 94.3 & 95.4 & 96.4 & 89.4 & 96.6 & 98.2 & 97.2 & 97.8 & 95.5 & 81.0 & 94.0 & 96.3 & 97.9 & 95.1 & 83.7 & 91.4 & 91.3 & 90.5 & 82.8 & 85.2 & 84.3 & 92.1 & 90.2 & 91.2 & 88.3 & 90.4 & 92.5 & 89.6 & 87.1 & 91.1 & 78.2 & 91.8 & 88.0 & 91.4 \\
\hline
\textit{dMat. Avg.} & 91.5 & 93.6 & 75.6 & 87.8 & 97.4 & 93.1 & 95.5 & 95.2 & 96.0 & 90.2 & 94.9 & 98.1 & 97.4 & 98.1 & 93.1 & 84.0 & 90.9 & 96.4 & 98.0 & 95.6 & 88.7 & 91.8 & 92.7 & 92.5 & 89.3 & 88.7 & 88.5 & 92.9 & 92.5 & 93.0 & 86.2 & 90.5 & 92.3 & 89.2 & 86.0 & 87.8 & 79.3 & 91.4 & 88.5 & 90.3 \\
\textit{LogSumExp} & 93.3 & 96.0 & 76.7 & 89.7 & 97.7 & 94.0 & 95.7 & 95.8 & \textbf{96.7} & 90.4 & 96.7 & 98.3 & 97.4 & 98.0 & 95.7 & 83.5 & 92.7 & 97.0 & 98.1 & 95.6 & 87.2 & 92.4 & 92.9 & 92.5 & 87.0 & 88.1 & 87.4 & 93.5 & 92.5 & 93.2 & 89.3 & 91.0 & 92.9 & 90.8 & 88.8 & 91.9 & 80.4 & 92.2 & 89.4 & 91.5 \\
\textit{Sum Descriptors} & 93.2 & 94.9 & 77.8 & 89.1 & 97.7 & 94.7 & \textbf{96.4} & 96.1 & 96.4 & 91.4 & 96.8 & 98.5 & 97.9 & \textbf{98.9} & 95.6 & 87.0 & 93.4 & \textbf{97.2} & \textbf{98.7} & 96.9 & 89.7 & 92.5 & 93.1 & 93.5 & \textbf{90.0} & 90.3 & \textbf{90.1} & 93.7 & 92.8 & 93.7 & 89.4 & 91.4 & 93.0 & 90.5 & 88.5 & 90.8 & 82.0 & 92.6 & 90.5 & 91.5 \\
\hline
\textit{\textbf{QR-FullVP (Ours)}} & \textbf{94.3 }& \textbf{96.5} & \textbf{79.4} & \textbf{90.6} & \textbf{98.0} & \textbf{95.1} & 96.3 & \textbf{96.2} & \textbf{96.7} & \textbf{91.5} & \textbf{97.2} & \textbf{98.6} & \textbf{98.0} & 98.8 & \textbf{96.3} & \textbf{87.6} & \textbf{94.5} & \textbf{97.2} & \textbf{98.7} & \textbf{97.1} & \textbf{89.5} & \textbf{93.1} & \textbf{93.6} & \textbf{93.7} & 89.3 & \textbf{90.4} & 89.5 & \textbf{94.1} & \textbf{93.0} & \textbf{93.8} & \textbf{91.4} & \textbf{91.9} & \textbf{93.7} & \textbf{91.5} & \textbf{90.4} & \textbf{93.6} & \textbf{83.8} & \textbf{92.9} & \textbf{90.6} & \textbf{92.3} \\
\hline
\end{tabular}
\vspace*{-1em}
\end{table*}
\begin{table}[!t]
  \centering
  \footnotesize
  \setlength{\tabcolsep}{3.5pt}
  \renewcommand{\arraystretch}{1.2}
  \caption{Recall@1 comparison of our QR approach, creating one place representation (FullVP) vs multiple representations using pairwise adjacent viewpoints (2VP), on SotonMV and Pitts250k.}
  \label{tab:2vp}
  \vspace{-1em}
  \subfloat{%
    \centering
    \begin{tabular}{|l|cccc|}
      \hline
      \textbf{Approach} & \multicolumn{4}{c|}{\textbf{SotonMV}} \\
      \cline{2-5}
      & \textbf{CosPlace} & \textbf{Eigenplaces} & \textbf{MixVPR} & \textbf{NetVLAD} \\
      & (2048D) & (2048D) & (4096D) & (4096D) \\
      \hline\hline
      \textbf{QR-FullVP} & 43.7 & 80.7 & 76.3 & 40.3 \\
      \textbf{QR-2VP} & 44.3 & 80.7 & 79.0 & 41.7\\
      \hline
    \end{tabular}
  }
  \hfill
  \subfloat{%
    \centering
    \begin{tabular}{|l|cccc|}
      \hline
      \textbf{Approach} & \multicolumn{4}{c|}{\textbf{SotonMV}} \\
      \cline{2-5}
      & \textbf{MegaLoc} & \textbf{SALAD} & \textbf{CricaVPR} & \textbf{BoQ} \\
      & (8448D) & (8448D) & (10752D) & (12288D) \\
      \hline\hline
      \textbf{QR-FullVP} & 81.7 & 81.3 & 60.3 & 87.7 \\
      \textbf{QR-2VP} & 82.7 & 82.3 & 62.0 & 88.7 \\
      \hline
    \end{tabular}
  }
    \hfill
  \subfloat{%
    \centering
    \begin{tabular}{|l|cccc|}
      \hline
      \textbf{Approach} & \multicolumn{4}{c|}{\textbf{Pitts250k-Test}} \\
      \cline{2-5}
      & \textbf{CosPlace} & \textbf{Eigenplaces} & \textbf{MixVPR} & \textbf{NetVLAD} \\
      & (2048D) & (2048D) & (4096D) & (4096D) \\
      \hline\hline
      \textbf{QR-FullVP} & 82.7 & 93.5 & 91.4 & 60.3 \\
      \textbf{QR-2VP} & 83.5 & 93.7 & 92.0 & 60.5 \\
      \hline
    \end{tabular}
  }
    \hfill
    \subfloat{%
    \centering
    \begin{tabular}{|l|cccc|}
      \hline
      \textbf{Approach} & \multicolumn{4}{c|}{\textbf{Pitts250k-Test}} \\
      \cline{2-5}
      & \textbf{MegaLoc} & \textbf{SALAD} & \textbf{CricaVPR} & \textbf{BoQ} \\
      & (8448D) & (8448D) & (10752D) & (12288D) \\
      \hline\hline
      \textbf{QR-FullVP} & 95.6 & 93.7 & 82.3 & 97.0 \\
      \textbf{QR-2VP} & 96.1 & 93.4 & 83.1 & 97.1 \\
      \hline
    \end{tabular}
  }

  \vspace{-2em}
\end{table}

\begin{table}[!ht]
\footnotesize
\centering
\setlength{\tabcolsep}{3.5pt} 
\renewcommand{\arraystretch}{1.2}

\caption{Recall@1 using query images aligned with reference map headings. Despite the additional non-overlapping viewpoint information, our approach suffers little to no degradation.}
\label{tab:aligned}
\vspace{-1em}
\subfloat{
\centering
\begin{tabular}{|l|cccc|}
\hline
\textbf{Approach} & \multicolumn{4}{c|}{\textbf{SotonMV}} \\ \cline{2-5} 
& \textbf{CosPlace} & \textbf{Eigenplaces} &
\textbf{MixVPR} & \textbf{NetVLAD} \\
& (2048D) & (2048D) & (4096D) & (4096D) \\
\hline\hline
\textit{Pooling} & 93.0 & 93.7 & 94.0 & 77.0 \\
\hline
\textit{dMat. Avg.}  & 65.0 & 77.3 & 83.3 & 50.3 \\
\textit{LogSumExp} & 93.3 & 93.3 & 93.0 & 76.0  \\
\textit{Sum Descriptors} & 78.3 & 84.7 & 84.3 & 59.3 \\
\hline
\textbf{QR-FullVP (Ours)} & 92.7 & 93.7 & 93.3 & 76.0 \\
\hline
\end{tabular}
}
\hfill
\subfloat{
\centering
\begin{tabular}{|l|cccc|}
\hline
\textbf{Approach} & \multicolumn{4}{c|}{\textbf{SotonMV}} \\ \cline{2-5} 
& \textbf{MegaLoc} & \textbf{SALAD} &
\textbf{CricaVPR} & \textbf{BoQ} \\
& (8448D) & (8448D) & (10752D) & (12288D) \\
\hline\hline
\textit{Pooling} & 95.0 & 95.7 & 94.7 & 95.7 \\
\hline
\textit{dMat. Avg.}  & 86.3 & 84.0 & 74.7 & 92.3 \\
\textit{LogSumExp} & 95.0 & 95.3 & 93.7 & 95.3 \\
\textit{Sum Descriptors} & 89.3 & 87.0 & 79.7 & 93.0 \\
\hline
\textbf{QR-FullVP (Ours)} & 95.0 & 94.0 & 94.0 & 95.3  \\
\hline
\end{tabular}
}
\hfill
\subfloat{
\centering
\begin{tabular}{|l|cccc|}
\hline
\textbf{Approach} & \multicolumn{4}{c|}{\textbf{Pitts250k-Test}} \\ \cline{2-5}
& \textbf{CosPlace} & \textbf{Eigenplaces} &
\textbf{MixVPR} & \textbf{NetVLAD} \\
& (2048D) & (2048D) & (4096D) & (4096D) \\
\hline\hline
\textit{Pooling} & 91.8 & 93.0 & 95.3 & 71.4 \\
\hline
\textit{dMat. Avg.}  & 64.0 & 77.4 & 84.4 & 42.8 \\
\textit{LogSumExp} & 92.2 & 93.3 & 94.3 & 69.5  \\
\textit{Sum Descriptors} & 72.4 & 82.1 & 87.5 & 48.5 \\
\hline
\textbf{QR-FullVP (Ours)} & 92.9 & 94.0 & 94.9 & 72.2 \\
\hline
\end{tabular}
} \hfill
\subfloat{
\centering
\begin{tabular}{|l|cccc|}
\hline
\textbf{Approach} & \multicolumn{4}{c|}{\textbf{Pitts250k-Test}} \\ \cline{2-5}
& \textbf{MegaLoc} & \textbf{SALAD} &
\textbf{CricaVPR} & \textbf{BoQ} \\
& (8448D) & (8448D) & (10752D) & (12288D) \\
\hline\hline
\textit{Pooling} & 98.1 & 97.0 & 95.8 & 97.8 \\
\hline
\textit{dMat. Avg.}  & 93.1 & 86.9 & 77.4 & 96.4 \\
\textit{LogSumExp} & 98.4 & 96.8 & 95.6 & 98.5 \\
\textit{Sum Descriptors} & 94.7 & 91.6 & 82.5 & 96.9 \\
\hline
\textbf{QR-FullVP (Ours)} & 98.7 & 97.3 & 95.9 & 98.7 \\
\hline
\end{tabular}
}

\vspace{-1em}
\end{table}

\subsection{Ablation Studies}

\textbf{Multiple Subspaces.} Building upon the decomposition of all images in a place, we evaluate the use of place representations using one or multiple subsets of available reference sets. Specifically, we evaluate on Nordland with CNN-based models and Oxford RobotCar with transformer-based models using seasonal-pair and weather-pair combinations in the reference rather than all others available. Further, we decompose places in SotonMV and Pitts250k-Test into multiple subspaces constructed using adjacent pairwise viewpoints, rather than all viewpoints together. 
With the Nordland reference pairs (see \tab{tab:nordlandpairs}), our QR approach outperforms other multi-reference approaches in the majority of cases (26/36), while outperforming pooling in all cases with significant gains, reflecting performance from using all reference images. Likewise with Oxford RobotCar reference pairs (\tab{tab:oxfordpairs}), our approach outperforms other multi-reference approaches in 36/40 cases. Despite continued improvement over single-reference set approaches, query conditions like Nordland Winter and Oxford RobotCar Night remain particularly challenging, where the top performing approach can occasionally vary with the reference sets used.

Deriving multiple subspaces from multi-view datasets (see \tab{tab:2vp}) yields, at best, marginal improvements over QR-FullVP, while introducing considerably greater computational overhead. The finer subspace decomposition on pairwise viewpoints allows queries to be matched with more specific and relevant subspaces, removing the influence of unrelated viewpoints that may harm discriminativeness. However, the results indicate that this benefit is minor. In contrast, the computational cost increases substantially, as additional decompositions are required, map storage grows from reused descriptors, and more comparisons are made during matching. In experimentation, we found that this additional overhead was not justified by the limited performance gain. A fully decomposed approach, which models all images collectively, offers a simpler, more general, and effective solution that considers all available information with minimal overhead.

\textbf{Aligned Viewpoints.} While previously we evaluate on multi-view reference collections with queries at intermediate headings, we investigate the robustness when queries are aligned with the reference headings and the task reduces to handling appearance variation. The results in \tab{tab:aligned} show that our decomposition approach maintains high robustness, achieving competitive accuracy to pooling despite integrating multiple headings with disjoint viewpoints. In contrast, score averaging and descriptor summing continue to show vulnerability, whose direct aggregation of irrelevant viewpoint descriptors leads to substantially reduced accuracy.

\section{Conclusion}
\label{sec:conclusion}
We address multi-reference VPR, introducing a descriptor-level mapping and matching framework that jointly models multiple reference observations with varying conditions via matrix decomposition into basis representations. Across appearance-change, viewpoint-change and datasets with unconstrained references, the method frequently improves Recall@1 over baseline multi-reference strategies while maintaining low computational cost. Our method is lightweight, with overhead further reduced through rank truncation and descriptor size compression. Overall, this formulation offers a practical and general mechanism for handling real-world variability in localization systems.

\textbf{Limitations and Future Work.} Our approach is most effective when references collectively span the appearance variation encountered at query time. However, like single-reference matching, performance degrades toward nearest-view matching when view coverage is sparse. Likewise, appearance conditions absent from the reference collections are poorly captured by the basis, resulting in large projection residuals. Therefore, fully representing possible query variations requires deliberate, extensive coverage under different conditions at collection time. Future work could also address residual gaps through learnable fusion techniques, non-linear subspace matching or semantic cues to relate disjoint observations.
Dense maps with heavy overlap contribute redundant information to the subspace while increasing the size of representations, impacting efficiency. This could be addressed at collection time or be identified prior to mapping via descriptor clustering or inspecting the descriptor matrix rank for dependencies. Future work could also investigate principled reference selection methods to maintain compact and informative subspaces. Finally, the framework assumes a fixed descriptor backbone. Maps constructed across different platforms or time with heterogeneous descriptors motivate future work on cross-descriptor subspace compatibility.






\bibliographystyle{IEEEtran} 
\bibliography{IEEEabrv,references}

\begin{IEEEbiography}[{\includegraphics[width=1in,height=1.25in,clip,keepaspectratio]{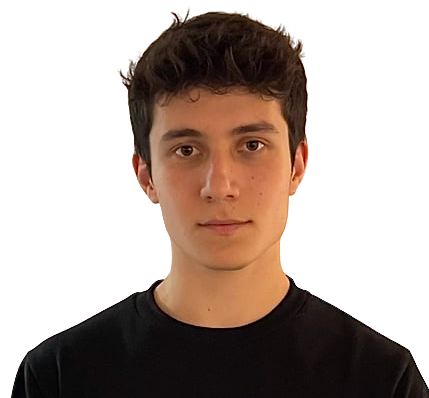}}]{Timur Ismagilov}
(Graduate Student Member, IEEE) received the M.Eng. degree in Electronic and Information Engineering, with First Class Honours, from Imperial College London, London, UK, in 2024. He is currently working toward the Ph.D. degree in Computer Science with the Agents, Interaction and Complexity Research Group at the University of Southampton, Southampton, UK. In affiliation with Responsible AI UK, he works on robust and efficient localization for robotics and mobile devices.
\end{IEEEbiography}

\begin{IEEEbiography}[{\includegraphics[width=1in,height=1.25in,clip,keepaspectratio]{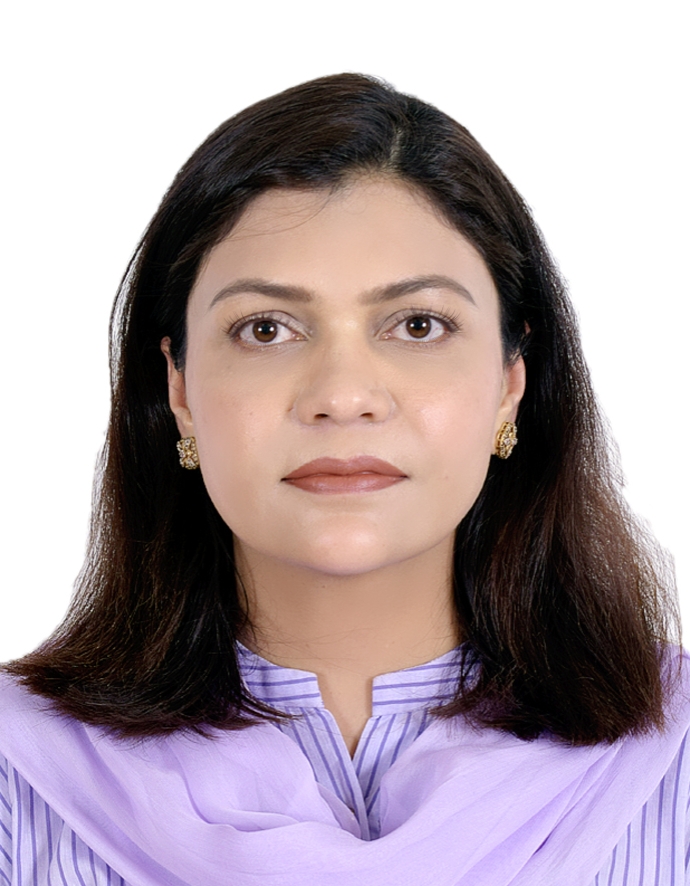}}]{Shakaiba Majeed}
received the B.Sc. degree in electrical engineering degree from the University of Engineering and Technology (UET) Taxila, Pakistan, in
2004, and the M.S. degree in electronic engineering and the Ph.D. degree in computer and software
engineering from Hanyang University, Seoul, South
Korea, in 2010 and 2018, respectively.
She has more than 20 years of industry experience
in embedded systems and artificial intelligence. Her
research interests include robotics, artificial intelligence, and embedded systems.
\end{IEEEbiography}

\begin{IEEEbiography}[{\includegraphics[width=1in,height=1.25in,clip,keepaspectratio]{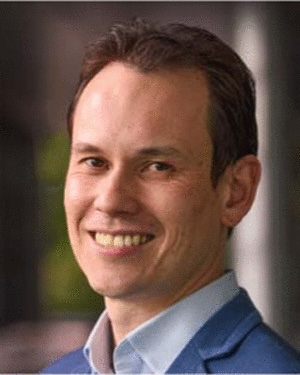}}]{Michael Milford}
 (Senior Member, IEEE) received
the bachelor’s degree in mechanical and space engineering and the Ph.D. degree in electrical engineering
from The University of Queensland, Brisbane, QLD,
Australia, in 2002 and 2006, respectively.
He is currently the Director with the QUT Centre
for Robotics, a Professor with the Queensland University of Technology, and is a Microsoft Research
Faculty Fellow. His research interests include the
neural mechanisms in the brain underlying tasks,
such as navigation and perception, to develop new
technologies in challenging application domains, such as all-weather, anytime
positioning for autonomous vehicles.
Dr. Milford is a Fellow of the Australian Academy of Technology and
Engineering and an Australian Research Council Laureate Fellow.
\end{IEEEbiography}

\begin{IEEEbiography}[{\includegraphics[width=1in,height=1.25in,clip,keepaspectratio]{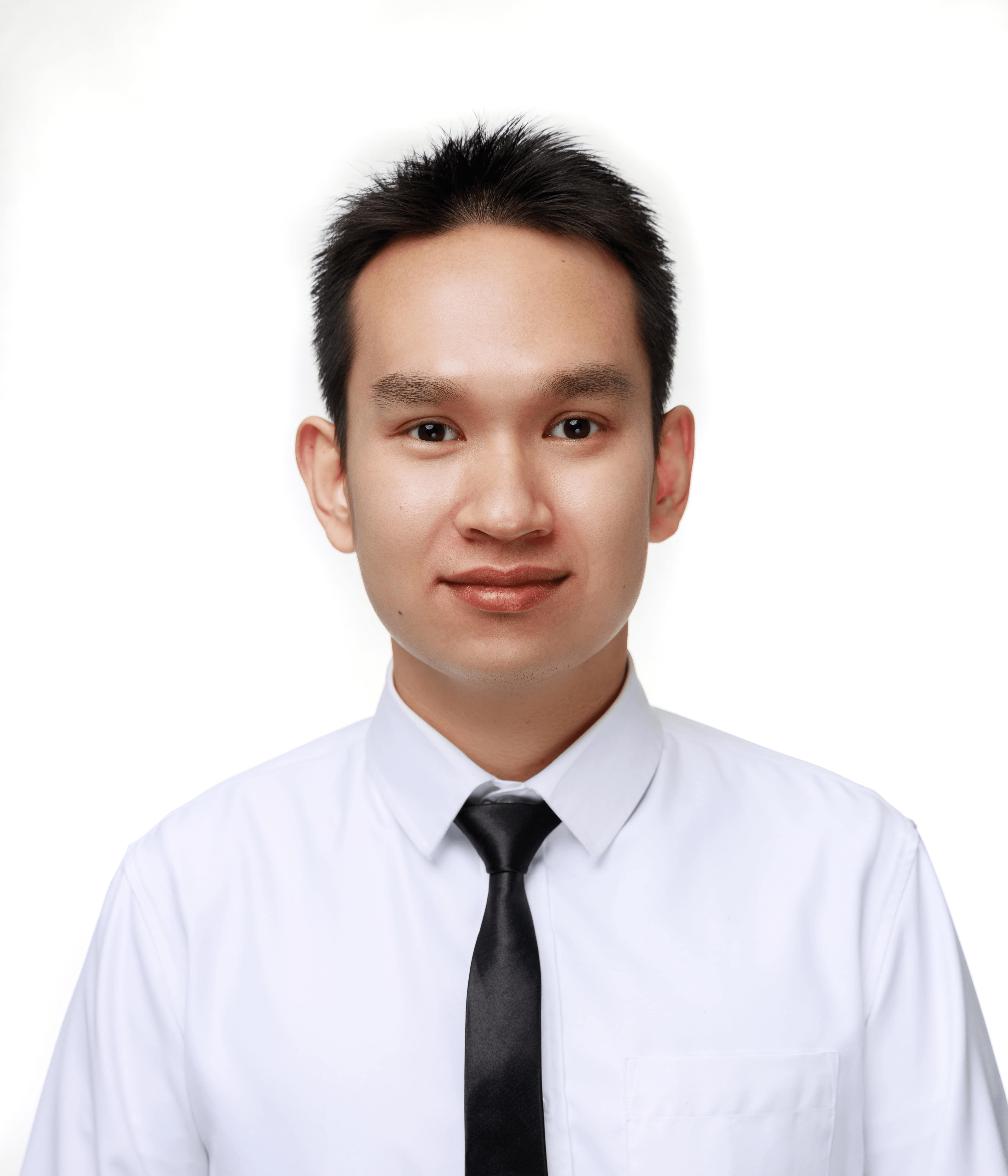}}]{Nguyen Tan Viet Tuyen}
(Member, IEEE) received his Ph.D. in Information Science from the Japan Advanced Institute of Science and Technology, Japan, in 2021. From 2021 to 2023, he was a Postdoctoral Research Associate at King’s College London, UK. Since 2023, he has been a New Frontiers Fellow (Lecturer) in AI at the School of Electronics and Computer Science, University of Southampton, UK. His research focuses on robotics, multimodal LLMs, and human-centred AI systems, with applications in human–AI and human–robot interaction.
\end{IEEEbiography}

\begin{IEEEbiography}[{\includegraphics[width=1in,height=1.25in,clip,keepaspectratio]{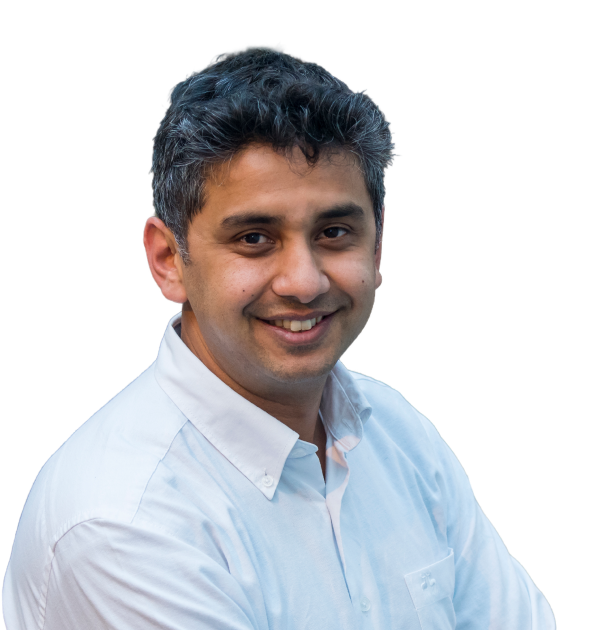}}]{Sarvapali D. Ramchurn}
received the Ph.D. degree in multi-agent systems from the University of Southampton, Southampton, U.K., in 2004. He is currently a Professor of Artificial Intelligence with the School of Electronics and Computer Science, University of Southampton, and a Fellow of the Institution of Engineering and Technology. He is CEO of Responsible AI UK and was formerly Director of the UKRI Trustworthy Autonomous Systems Hub. His research focus is human-AI collaborations, using techniques from Machine Learning, HCI, and Game Theory, applied in areas including AI-based disaster response using multi-UAV systems and AI-driven large-scale battery management for smart grids. He was the recipient of the AXA Research Fund Award (2018) for his work on Responsible AI, and has received multiple best paper awards for his research. He has founded several startups, including Empati AI and Sentient Sports, and is a member of the UK HM Government's Digital Service Responsible AI Advisory Panel and the WHO Expert AI Advisory Panel.
\end{IEEEbiography}

\begin{IEEEbiography}[{\includegraphics[width=1in,height=1.25in,clip,keepaspectratio]{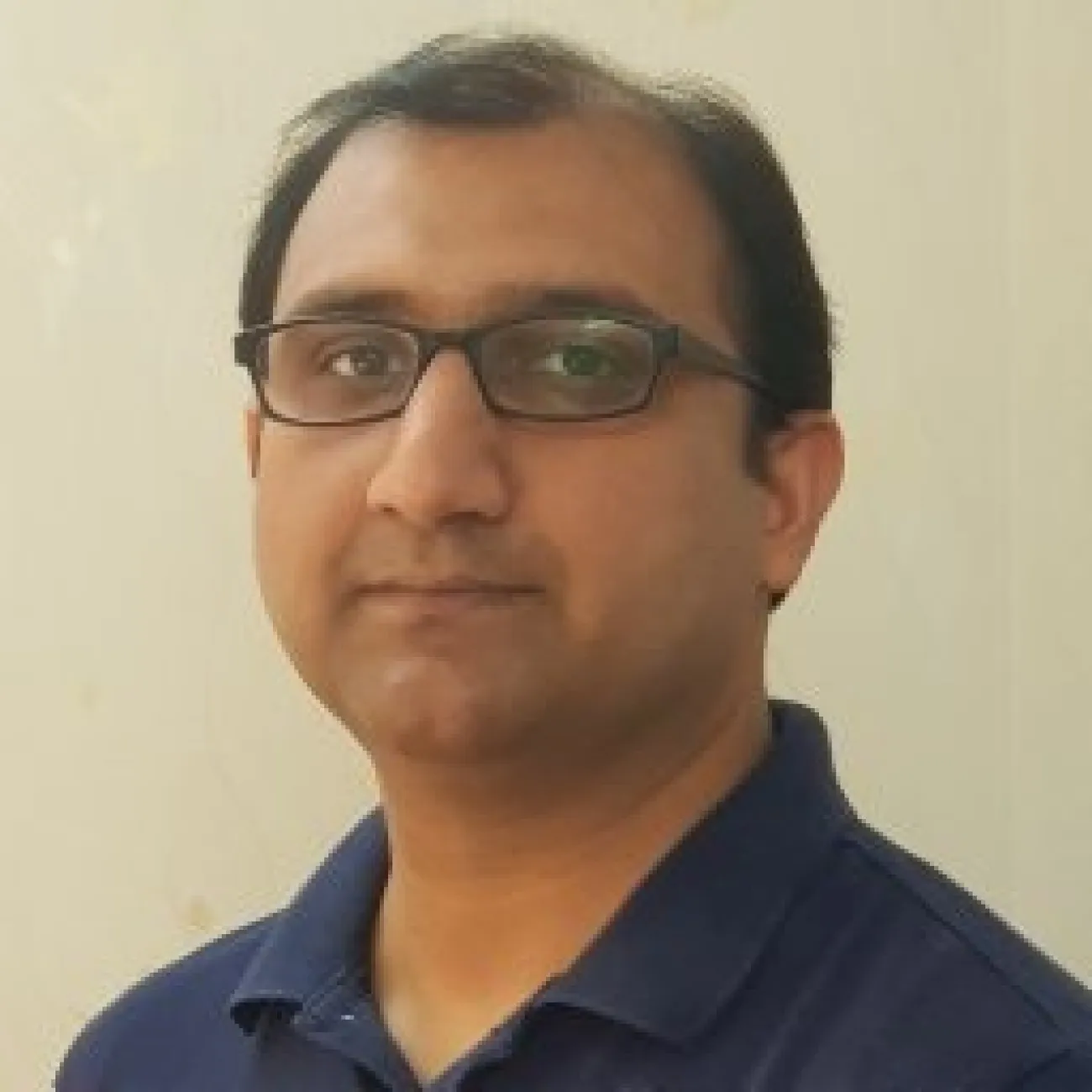}}]{Shoaib Ehsan} (Senior Member, IEEE)
 received the
B.Sc. degree in electrical engineering degree with
first-class honours from the University of Engineering
and Technology, Taxila, Pakistan, in 2003, and the
Ph.D. degree in computing and electronic systems
from the University of Essex, Colchester, U.K., in
2012.
He is currently an Associate Professor in AI
and robotics with the University of Southampton,
Southampton, U.K., and a part-time Reader in Spatial
AI for Robotics and Autonomous Systems with the
University of Essex. His research interests include robot vision, autonomous
systems, and resource-efficient AI.
Dr. Ehsan was the recipient of the Royal Academy of Engineering as a
Researcher with Exceptional Promise, the 2024 IEEE Robotics and Automation
Letters Best Paper Award - Honourable Mention, and the British Machine Vision
Association’s Sullivan Doctoral Thesis Prize for the best U.K. Ph.D. thesis
in computer vision. He was an Associate Editor for IEEE TRANSACTIONS ON
ROBOTICS and IEEE ROBOTICS AND AUTOMATION LETTERS.
\end{IEEEbiography}

\end{document}